\documentclass{article} 
\PassOptionsToPackage{table}{xcolor}
\usepackage{iclr2027_conference,times}
\iclrpreprintcopy
\usepackage{amsmath,amsfonts,bm}

\def\eqref#1{equation~\ref{#1}}

\def\1{\bm{1}}

\DeclareMathAlphabet{\mathsfit}{\encodingdefault}{\sfdefault}{m}{sl}
\SetMathAlphabet{\mathsfit}{bold}{\encodingdefault}{\sfdefault}{bx}{n}

\usepackage{amsmath}
\usepackage[utf8]{inputenc} 
\usepackage[T1]{fontenc}    
\usepackage{hyperref}       
\hypersetup{hidelinks}
\usepackage{url}            
\usepackage{booktabs}       
\usepackage{amsfonts}       
\usepackage{nicefrac}       
\usepackage{microtype}      
\usepackage{xcolor}         
\usepackage{graphicx}
\usepackage{multirow}

\usepackage{wrapfig}
\usepackage{placeins}
\usepackage{array}
\usepackage{pifont}
\usepackage{caption}
\usepackage{subcaption}

\usepackage{titlesec,enumitem}

\titlespacing*{\section}{0pt}{0.7ex}{0.3ex}
\titlespacing*{\subsection}{0pt}{0.6ex}{0.2ex}
\titlespacing*{\subsubsection}{0pt}{0.4ex}{0.2ex}

\setlist{nosep,leftmargin=*}

\title{\textsc{MolWorld}: Molecule World Models for Actionable Molecular Optimization}

\author{%
  Yang Qiao\thanks{Equal contribution.} \\
  Emory University \\
  Atlanta, GA, USA \\
  \texttt{yqiao47@emory.edu} \\
  \And
  Bo Pan\footnotemark[1] \\
  Emory University \\
  Atlanta, GA, USA \\
  \texttt{bo.pan@emory.edu} \\
  \And
  Hao-Wei Pang \\
  Merck \& Co., Inc. \\
  Rahway, NJ, USA \\
  \texttt{hao-wei.pang@merck.com} \\
  \AND
  Peter Zhiping Zhang \\
  Merck \& Co., Inc. \\
  Rahway, NJ, USA \\
  \texttt{zhiping.peter.zhang@merck.com} \\
  \And
  Liying Zhang \\
  Merck \& Co., Inc. \\
  Rahway, NJ, USA \\
  \texttt{liying.zhang@merck.com} \\
  \And
  Liang Zhao \\
  Emory University \\
  Atlanta, GA, USA \\
  \texttt{liang.zhao@emory.edu} \\
}
\begin{document}

\maketitle

\begin{abstract}
Molecular optimization in drug discovery aims to discover molecules with improved
target properties, but practical lead optimization often requires more than high
predicted scores. A useful candidate should also be actionable: it should be
reachable from known molecules through a sequence of local structural
modifications, providing explicit structural references for interpreting property
changes within an evolving chemical series. Existing de novo and single-molecule optimization methods do not
explicitly model such reachability, especially when both the target molecules and
the intermediate molecules connecting them to known compounds are unknown. In
this work, we formulate actionable molecular optimization as iterative expansion
of a molecule-transfer graph, where nodes are molecules and edges encode matched
molecular pair (MMP) relations representing localized structural differences. We propose \textsc{MolWorld}, a molecule world
model-guided framework that uses connected local subgraphs as anchor contexts for
graph expansion. Given the current graph and a candidate context, a latent molecule
world model predicts the resulting local structural expansion. Its dynamics are
shared across property objectives, while a property-specific scorer evaluates
predicted expansions to guide context selection. A context-conditioned generator
then uses the molecular structures and MMP relations within selected contexts to
propose candidates, learning from local graph-completion tasks. Candidates are
evaluated, and those connected through verified MMP relations are incorporated
into the graph, preserving reachability from the initial set. The updated graph serves as the state for
subsequent optimization.
Experiments on property optimization and
docking-based tasks show that \textsc{MolWorld} discovers high-property molecules
while maintaining stronger structural connectivity, supporting
actionable and sequential molecular design.
\end{abstract}

\section{Introduction}

\begin{wrapfigure}{R}{0.45\columnwidth}
    \centering
    \includegraphics[width=0.4\columnwidth]{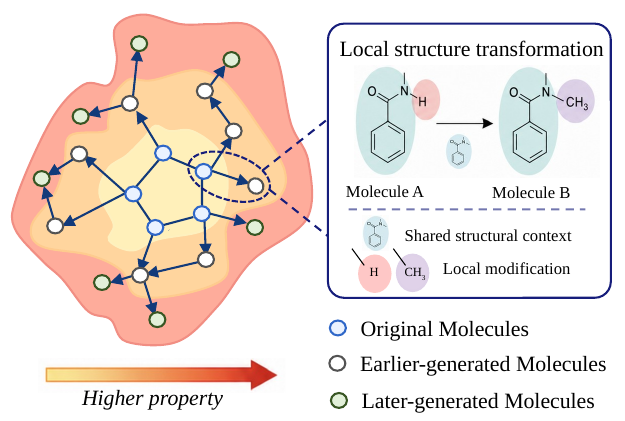}
    \caption{Molecule-transfer graph expansion. Generated molecules are progressively inserted into the graph through verified MMP relations, enabling sequential exploration toward higher-property regions while preserving reachability from the initial molecule series.}
    \label{fig:intro}
\end{wrapfigure}

Molecular optimization aims to design molecules with improved target properties,
such as activity, selectivity, docking affinity, or drug-likeness~\cite{nicolaou2007molecular}. Recent machine
learning methods have made substantial progress by formulating this problem as
de novo molecular generation \cite{kusner2017grammar,liu2018constrained, jin2018junction, zhou2019optimization, guimaraes2017objective, you2018graph} or property-guided optimization from existing
molecules \cite{he2021molecular,he2022transformer,jin2020hierarchical,fu2020core, wu2024leveraging}. These formulations are effective for discovering high-scoring
candidates under a property oracle, but practical lead optimization often requires
more than high predicted property values. A useful candidate should also be
\emph{actionable}: it should be reachable from existing molecules through a sequence
of local structural modifications, providing explicit structural references for
interpreting property changes and planning further exploration within a chemical series.

In lead
optimization, molecules are commonly organized around analogue series, where
compounds share conserved structural contexts and differ through local
substitutions or modifications~\cite{naveja2021automatic, stacey2025visualising, raymond2009rationalizing, yoshimori2022deepas}. Such relationships naturally induce a
\emph{molecule-transfer graph}: nodes are molecules, and edges represent matched
molecular pair (MMP) relations~\cite{dossetter2013matched, griffen2011matched}. An MMP consists of two molecules that retain the
same structural context while differing by a well-defined, localized fragment
replacement, as illustrated in the
inset of Figure~\ref{fig:intro}. Each edge therefore provides an explicit structural
comparison for interpreting property differences, while successive edges make the
progression from existing compounds to new candidates traceable. Such a sequence
can extend beyond the immediate structural neighborhood of any single starting molecule.


We therefore study \emph{actionable molecular optimization}: as illustrated in
Figure~\ref{fig:intro}, given an initial set
of related molecules, the goal is to discover molecules with strong properties while ensuring that they are reachable from the initial set through
paths of MMP relations.
After a new molecule is generated and connected to the current series, it becomes
part of the molecular world from which future molecules can be designed.
Optimization therefore proceeds through iterative expansion of this molecular
world, with the updated graph serving as the state for subsequent exploration.

Achieving this goal presents three challenges. First, the existing graph contains
rich information about molecular structures and local transformations, but its
size makes directly conditioning generation on the entire graph impractical. A
generation context must retain useful structural relationships while remaining
manageable as the graph grows. Second, the graph contains many potential expansion
regions, and the properties of existing molecules do not directly reveal the
quality of new molecules that can be generated from them. Evaluating every
possible expansion would require costly generation and property-oracle calls.
Third, generating candidates that can extend the graph requires exploiting both
the structures of related molecules and the MMP relations linking them. The generator must use these patterns to propose candidates connected to the current graph.

Motivated by this view, we propose \textsc{MolWorld}, a molecule world
model-guided framework for actionable molecular optimization. We use connected
local subgraphs, termed anchor contexts, to provide compact generation contexts
that retain related molecular structures and their MMP relations. Building on this
representation, \textsc{MolWorld} couples two modules. First, molecule world
model-guided anchor context selection anticipates the outcomes of candidate
expansions before generation is executed. A latent molecule world model predicts
the local structural expansion associated with each candidate context, and a
property-specific scorer combines this prediction with the current graph
representation to rank the contexts. The structural dynamics model is shared
across property objectives. Second, a context-conditioned molecule generator
jointly uses the molecular structures and MMP relations within a selected context
to propose candidates that complete or extend its local molecule-transfer pattern.
Candidates are evaluated, and those connected through verified MMP relations are
incorporated into the graph. The updated graph serves as the state for the next iteration.

We evaluate \textsc{MolWorld} on molecular property optimization and
docking-based optimization tasks. The results show that \textsc{MolWorld}
discovers high-property molecules while maintaining stronger graph
connectivity than competing methods.
Our contributions are summarized as follows:
\begin{itemize}
    \item We formulate actionable molecular optimization as a reachability-constrained
    problem and propose \textsc{MolWorld}, a framework that optimizes molecular
    properties through iterative expansion of a molecule-transfer graph. By
    maintaining MMP paths to the initial molecular set, the framework makes
    high-property candidates traceable to the starting molecular series and provides
    structural references for interpreting property changes along local modifications.

    \item We introduce a molecule world model that simulates the evolution of a
    molecular world under context-conditioned generation. Given the current
    molecule-transfer graph and an anchor context, it anticipates the resulting
    structural expansion in latent space. Coupled with a property-specific scorer,
    it guides property-directed search by evaluating expansions before execution.

    \item We formulate context-conditioned molecular generation as local
    molecule-transfer graph completion, using multiple related molecules and their
    MMP relations as a joint generation context. The generator learns to propose
    candidates from the structural patterns represented by this context, supporting
    graph expansion through subsequent MMP validation.

\end{itemize}

\section{Related Work}
\textbf{Machine Learning-based Molecular Optimization.}
\textbf{De novo molecular optimization} generates molecules with desired properties without a specified starting compound. Approaches include continuous latent-space optimization \cite{kusner2017grammar,liu2018constrained, jin2018junction}, reinforcement learning \cite{zhou2019optimization, guimaraes2017objective, you2018graph, olivecrona2017molecular, munson2024denovo, wang2025efficient}; graph-based autoregressive or flow models \cite{shi2020graphaf, zang2020moflow}; and diffusion and structure-aware models that generate 2D or 3D molecules under geometric or target-conditioned constraints \cite{hoogeboom2022equivariant,xu2023geometric,peng2022pocket2mol,guan2023targetdiff,schneuing2024structure, huang2024dual}. In contrast, \textbf{pairwise molecular optimization} seeks improved analogs of a lead molecule while preserving similarity, scaffolds, or key substructures. Approaches include molecular translation using paired textual representations \cite{he2021molecular,he2022transformer}, graph-to-graph translation \citep{jin2019multimodal,jin2019hierarchical,jin2020hierarchical,fu2020core,xie2021mars}, and LLM-based approaches \cite{wu2024leveraging, ye2025drugassist}. Retrieval-based methods such as RetMol~\citep{wang2022retrieval} also condition generation on multiple exemplar molecules. \textsc{MolWorld} instead organizes molecular contexts within an evolving MMP graph and uses a molecule world model to predict graph expansion for context selection.

Additional related work on World Models is given in Appendix~\ref {app:additional_related_work}.

\section{Method}
\label{sec:method}

\begin{figure}[t]
  \centering
  \includegraphics[width=\textwidth]{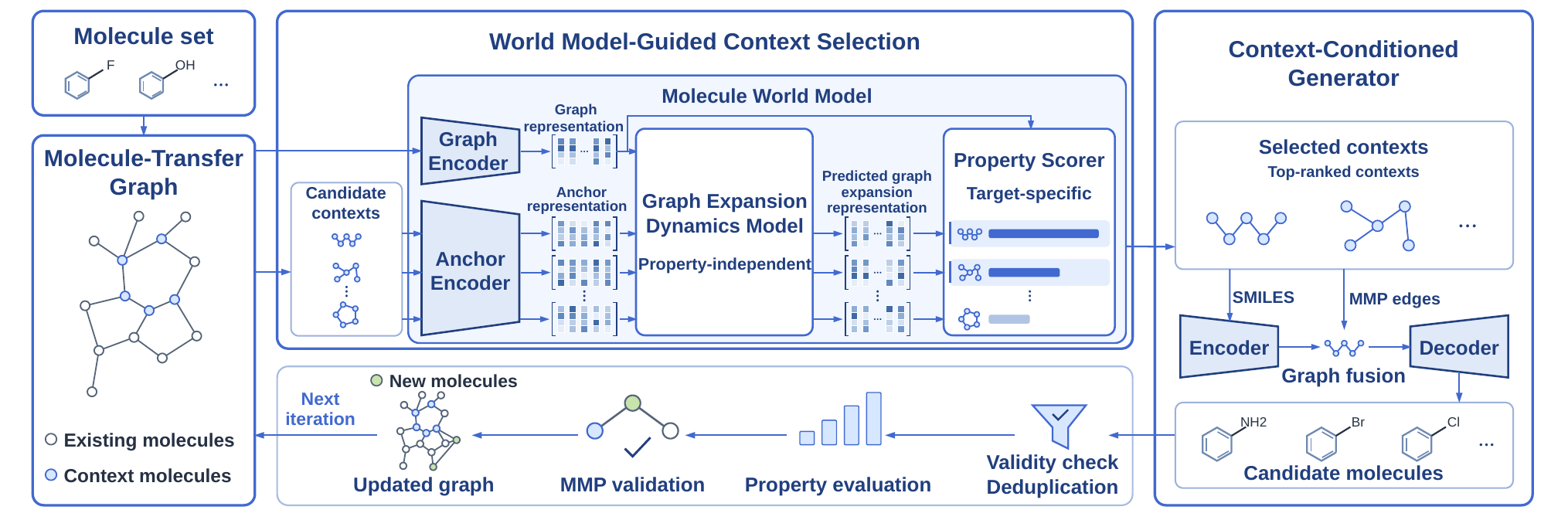}
    \caption{Overview of the \textsc{MolWorld} framework. The molecule world model predicts graph expansion conditioned on candidate anchor contexts, and a property-specific scorer ranks these contexts for selection. The context-conditioned generator then proposes new molecules. After validity checks, deduplication, and property evaluation, candidates connected to the current graph through verified MMP relations are incorporated for the next iteration.}  \label{fig:main}
\end{figure}

Given an initial molecular set and a target property, we propose \textsc{MolWorld},
a framework for actionable molecular optimization through iterative expansion of a
molecule-transfer graph. The framework generates high-property candidates while
maintaining MMP paths to the initial set, and incorporates accepted molecules into
the graph for subsequent optimization. Within this framework, a molecule world
model predicts how generation from each candidate context would expand the graph,
guiding anchor context selection. A context-conditioned generator then uses the
molecular structures and MMP relations within the selected contexts to propose
new molecules (Figure~\ref{fig:main}).

\subsection{Problem Formulation}

Given an initial set of related molecules $\mathcal{M}_0$ and a target property
oracle $f$, we seek new molecules with high property values that can be reached
from $\mathcal{M}_0$ through successive local structural modifications. We use
matched molecular pair (MMP) relations to specify these modifications:
$m\sim m'$ indicates that $m'$ can be obtained from $m$ by a localized structural
change that preserves their shared chemical context. A molecule $x$ is
\emph{reachable} from $\mathcal{M}_0$ if there exists a finite transformation path
\begin{equation}
m_0\rightarrow m_1\rightarrow\cdots\rightarrow m_L=x,
\qquad m_0\in\mathcal{M}_0,
\label{eq:reachability}
\end{equation}
where $m_{\ell-1}\sim m_\ell$ for every $\ell=1,\ldots,L$. Each generated molecule
must be connected to the initial set through a path of MMP relations among initial
and generated molecules. Our objective is to find a generated set $\mathcal{X}$
that maximizes
\begin{equation}
\max_{\mathcal{X}}\;J_f(\mathcal{X})
\label{eq:reachable_optimization}
\end{equation}
subject to the reachability requirement above, where $J_f$ measures the property
quality of the generated set, such as its top-$k$ property scores. This formulation
supports both direct analogues of initial molecules and more distant candidates
connected through successive local modifications.

\subsection{Overall Framework}
\label{sec:framework}

\textbf{Graph and anchor contexts.}
We initialize the molecule-transfer graph as
$G_0=(\mathcal{V}_0,\mathcal{E}_0)$, where $\mathcal{V}_0=\mathcal{M}_0$ and edges
encode MMP relations between molecules. We denote the graph at iteration $t$ by
$G_t=(\mathcal{V}_t,\mathcal{E}_t)$. Since the graph can already contain many
molecules and continues to grow during optimization, directly conditioning
molecular generation on the entire graph becomes computationally impractical.
We therefore condition generation on a connected local subgraph $Z=(V_Z,E_Z)$
of $G_t$, termed an \emph{anchor context}. The molecular structures and MMP relations within $Z$ capture shared structural features and local transformation patterns, providing a compact generation context. Each anchor context contains five connected molecules. The framework does not require a fixed initial graph size: any initial graph containing a connected component of at least five molecules can supply a valid generation context.

\textbf{World model-guided context selection.}
At iteration $t$, we first construct a candidate pool $\mathcal{C}_t$ using
connected-subgraph beam search, combining high-property seeds and neighbors with
randomized proposals favoring less-visited molecules. Repeated-use penalties
discourage revisiting the same contexts, balancing property-guided search with
exploration. Construction details are provided in Appendix~\ref{app:anchor_selection}.
Given the current graph $G_t$ and a candidate anchor context $Z\in\mathcal{C}_t$,
the molecule world model, detailed in Section~\ref{sec:anchor_selection}, predicts
the graph expansion resulting from generation conditioned on $Z$. To assess this
expansion for the target property, its property-specific scorer combines the
prediction with the current graph representation to assign a score $s_t(Z)$.
For a generation batch size $B$, we select the $B$ highest-scoring contexts:
\begin{equation}
\label{eq:anchor_final_selection_main}
\mathcal{Z}_t
=\underset{\mathcal{Z}\subseteq\mathcal{C}_t,\;|\mathcal{Z}|=B}
{\arg\max}
\sum_{Z\in\mathcal{Z}}s_t(Z).
\end{equation}

\textbf{Generation and graph expansion.}
The context-conditioned generator (detailed in Section~\ref{sec:generator}) uses
the molecular structures and MMP relations within the selected contexts to
produce candidates $\mathcal{X}_t$. After validity checks and deduplication, all
remaining candidates are evaluated by the property oracle. RDKit-based MMP
matching identifies those connected to at least one molecule in $G_t$, forming
the accepted set $\widetilde{\mathcal{X}}_t$. The graph is then updated as
\begin{equation}
\label{eq:world_transition_main}
G_{t+1}
=\left(
\mathcal{V}_t\cup\widetilde{\mathcal{X}}_t,\,
\mathcal{E}_t\cup\mathcal{E}_t^{+}
\right),
\end{equation}
where $\mathcal{E}_t^{+}$ contains verified MMP edges linking accepted molecules
to existing or other accepted molecules. Each accepted molecule extends an
existing MMP path from the initial set. The updated graph supplies new anchor
contexts for subsequent generation, completing the feedback loop in
Figure~\ref{fig:main}. Detailed acceptance and edge-construction rules are
provided in Appendix~\ref{app:graph_update}.

\subsection{Molecule World Model}
\label{sec:anchor_selection}

The properties of molecules in an anchor context indicate promising regions of
the graph, but do not directly reveal the quality of new molecules that generation
from this context would produce. Evaluating every candidate context through
generation and property evaluation would be costly. Following latent world-model
approaches~\citep{ha2018worldmodels,hansen2024tdmpc2}, we learn to predict
action-conditioned changes to the search state in representation space. Here,
the state is the current molecule-transfer graph, an action is generation
conditioned on an anchor context, and the predicted outcome represents the
resulting structural expansion. Given the current
graph $G_t$ and a candidate anchor context $Z\in\mathcal{C}_t$, our molecule world
model combines graph expansion prediction and property scoring as follows:
\begin{equation}
\label{eq:anchor_latent_dynamics}
\begin{aligned}
\text{State and context encoders:}\quad
&h_t=E_G(G_t),\qquad a_Z=E_A(Z),\\
\text{Graph expansion dynamics model:}\quad
&\widehat\delta_Z=\mathcal{W}_\phi(h_t,a_Z),\\
\text{Property scorer:}\quad
&s_t(Z)=Q_\psi(h_t,\widehat\delta_Z).
\end{aligned}
\end{equation}
Here, $E_G$ encodes the complete current graph $G_t$ as $h_t$. Each anchor
context $Z$ specifies a context-conditioned generation action and is encoded by
$E_A$ as $a_Z$. Conditioned on both representations, the graph expansion dynamics model
$\mathcal{W}_\phi$ predicts how the graph would expand under this action,
representing the anticipated expansion in latent space as $\widehat\delta_Z$.
The scorer $Q_\psi$ combines this prediction with the current graph representation
to assign the context a score $s_t(Z)$. The dynamics model learns structural
expansion independently of the target property and is shared across property
objectives, while the scorer is trained for the target property $f$.

For the anchor encoder $E_A$, we reuse the molecular encoding and graph-fusion
modules of the context-conditioned generator, pretrained on a separate
ChEMBL-derived dataset as described in Section~\ref{sec:generator}. This connects
the world model's action representation to the context used for generation.
We train the remaining components of the molecule world model on an offline
dataset $\mathcal{D}$ of molecular graph expansions constructed from
ChEMBL~\cite{zdrazil2024chembl} using the pretrained generator. The dataset records
molecular graphs, anchor contexts, and their observed expansion outcomes; its
construction and train/validation/test splits are described in
Appendix~\ref{app:transition_dataset}.

To support graph expansion prediction, $E_G$ should capture both node-level
structural context and graph-level information. We therefore adapt masked latent
prediction~\cite{baevski2022data2vec,assran2023self} to the initial graphs in
$\mathcal{D}$, combining node and graph prediction with token-diversity
regularization. During offline pretraining, a student graph encoder and prediction
heads predict representations from graphs with masked node features, while an
exponential-moving-average (EMA) target encoder supplies representations of the
corresponding complete graphs. The objective is
\begin{equation}
\label{eq:graph_encoder_loss_main}
\mathcal{L}_{\mathrm{enc}}
=\frac{1}{|M|}\sum_{v\in M}
\ell_{\mathrm H}\!\left(\widehat u_v,\operatorname{sg}[\bar u_v]\right)
+\lambda_g\ell_{\mathrm H}\!\left(\widehat H,\operatorname{sg}[\bar H]\right)
+\lambda_d\mathcal{L}_{\mathrm{div}}.
\end{equation}
Here, $M$ is the set of masked molecular nodes; $\widehat u_v,\widehat H$ are
node and graph-token predictions, and $\bar u_v,\bar H$ are the corresponding
layer-normalized target representations. The function $\ell_{\mathrm H}$ is
smooth-$L_1$ loss, $\operatorname{sg}$ denotes stop-gradient, and
$\lambda_g,\lambda_d$ are loss weights. The diversity term
$\mathcal{L}_{\mathrm{div}}$ discourages redundant graph tokens, encouraging
complementary structural information. After pretraining, the EMA target encoder
is frozen and used as $E_G$. The graph encoder architecture, training details,
and validation results are provided in Appendix~\ref{app:world_model_details}.

With both encoders fixed, the dynamics model learns to predict the graph
expansion induced by generation conditioned on $Z$. Since the structural change
from a single generation action is small relative to the current graph, we focus
supervision on the added structure. The prediction target therefore encodes an
expansion graph comprising accepted new molecules, their MMP edges, and the
existing molecules to which they connect. The training objective is
\begin{equation}
\label{eq:anchor_dynamics_loss}
\mathcal{L}_{\mathrm{dyn}}(\phi)=\mathbb{E}_{(G_t,Z)\sim\mathcal{D}}
\left[\ell_{\mathrm H}\!\left(
\mathcal{W}_\phi\!\left(E_G(G_t),E_A(Z)\right),\;
\frac{1}{R_Z}\sum_{r=1}^{R_Z}E_G\!\left(\Delta G_{t,Z}^{(r)}\right)
\right)\right].
\end{equation}
Here, $\Delta G_{t,Z}^{(r)}$ is the $r$-th observed expansion graph, and $R_Z$ is
the number of recorded generations for the same state--context pair. The dynamics
model architecture, target construction, training details, and validation results
are provided in Appendix~\ref{app:world_model_details}.

Finally, we train the property-specific scorer $Q_\psi$ using the current graph
representations and the dynamics model's predicted expansion representations,
keeping both encoders and the dynamics model fixed. Supervision reflects the
property quality of the best newly accepted molecules. The training objective
combines a best-action loss that favors the highest-quality actions within each
training state with a pairwise ranking loss that learns their relative ordering.
Supervision and loss definitions are provided in Appendix~\ref{app:anchor_selection}.

\subsection{Context-Conditioned Molecule Generation}
\label{sec:generator}

Given a selected context $Z\in\mathcal{Z}_t$, the generator proposes candidates that locally
complete or extend the molecule-transfer pattern represented by $Z$. 
For an anchor context $Z=(V_Z,E_Z)$, the generator models a conditional distribution:
\begin{equation}
\label{eq:generator_conditional}
p_\theta(x \mid Z)
\approx
p_{\mathrm{data}}\big(x \mid x \in \mathcal{N}_{\mathrm{local}}(Z)\big),
\end{equation}
where $\mathcal{N}_{\mathrm{local}}(Z)$ denotes molecules that plausibly complete
or extend the local molecule-transfer graph represented by $Z$. This formulation
turns candidate generation from one-to-one molecular editing into local graph
completion: instead of modifying a single molecule in isolation, the model learns to propose molecules compatible with the transformation patterns in the selected context.

We use an encoder-decoder architecture to parameterize this distribution,
as illustrated in Figure~\ref{fig:main}.
Each molecule $m_i\in V_Z$ is represented by its SMILES string. A shared
molecular encoder first maps the molecules in the context into node
representations. A graph fusion module then performs message passing over the
context edges $E_Z$ to obtain a structure-aware context representation:
\begin{equation}
\label{eq:context_representation}
C_Z
=
\mathrm{Pool}
\left(
\left\{
\mathrm{GNN}_\theta
\big(
\{\mathrm{Enc}_\theta(\mathrm{SMILES}(m_i))\}_{m_i\in V_Z},
E_Z
\big)_i
\right\}_{m_i\in V_Z}
\right).
\end{equation}
Here, $C_Z$ encodes the molecular identities of the context nodes and their local transformation relations. The molecular encoding and graph-fusion modules are shared with the anchor encoder $E_A$: the generator uses the fused token sequences for decoding, while $E_A$ averages them within each molecule to obtain the action representation $a_Z$.
A Transformer decoder conditions on $C_Z$ for autoregressive SMILES generation:
\begin{equation}
\label{eq:generator_autoregressive}
p_\theta(x \mid Z)
=
\prod_{\ell=1}^{L}
p_\theta(x_\ell \mid x_{<\ell}, C_Z).
\end{equation}
Sampling from this distribution yields multiple candidate molecules conditioned
on the same selected context. The generator is trained in two stages, first using
pairwise molecular transformations to pretrain the Transformer encoder-decoder, then using
subgraph-based contexts to train all modules. Training details are provided in
Appendix~\ref{app:generator_details}.

\section{Experiments}

\subsection{Experimental Setup}

We pretrain the context-conditioned generator and the molecule world model on ChEMBL~\citep{zdrazil2024chembl},
which provides a large-scale collection of bioactive molecules for learning local molecular transformation patterns, 
and evaluate \textsc{MolWorld} on ZINC250~\citep{Sterling2015ZINC15} and PMV21~\citep{Vu2021p53}. 
For each dataset, we construct a molecule-transfer graph whose nodes are molecules and whose edges encode MMP relations identified using RDKit-based molecular fragmentation.
Implementation details and hyperparameters are provided in Appendix~\ref{app:implementation_details}.

\textbf{Task 1: Property Optimization.}
We adopt the oracle budgets during optimization and running top-$k$ AUC evaluation of the Practical Molecular Optimization (PMO) benchmark~\citep{gao2022sample} on both ZINC250 and PMV21.
Given a set of molecules, the task is to iteratively generate new molecules under a fixed oracle budget to improve a target property.
We consider four target properties: GSK3$\beta$ (glycogen synthase kinase-3 beta),
QED~\citep{bickerton2012quantifying}, which measures drug-likeness,
JNK3 (c-Jun N-terminal kinase-3), and DRD2 (dopamine receptor D2).
For the three bioactivity tasks, pretrained predictive models from TDC are used as oracle functions.
The oracle budget is set to 10,000 calls for ZINC250 and 1,000 calls for PMV21. 
Each experiment is repeated with three random seeds.

\textbf{Task 2: Docking-Based Optimization.}
We further consider three protein-ligand docking tasks from TDC \cite{graff2021accelerating}, 
which correspond to structure-based drug design scenarios. 
The proteins include DRD3 (PDB ID: 3PBL), EGFR (PDB ID: 2RGP), and Adenosine A2A receptor (PDB ID: 3EML). 
For each task, molecules are evaluated using the TDC docking oracle implemented through the PyScreener AutoDock Vina backend \cite{eberhardt2021autodock}. 
The oracle returns the best Vina binding affinity among generated poses, where lower docking scores indicate stronger predicted binding. 
All methods are evaluated under the same docking protocol and oracle budget. 
Due to the higher computational cost of docking evaluation, we restrict the oracle budget to 1,000 calls. 
The full docking protocol, including receptor preparation, ligand conformer generation, docking box settings, and Vina parameters, is provided in Appendix~\ref{app:docking_protocol}.


\textbf{Compared Methods.}
We compare \textsc{MolWorld} with nine baselines for property optimization: GraphGA~\citep{jensen2019graph}, REINVENT~\citep{olivecrona2017molecular}, Augmented Memory~\citep{guo2023augmented}, GP-BO~\citep{tripp2021fresh}, MOLLEO~\citep{wang2025efficient}, NFBO~\citep{lee2025latent}, GenMol~\citep{lee2025genmol}, RetMol~\citep{wang2022retrieval}, and InVirtuoGen~\citep{kaech2026refine}.
For docking-based optimization, we compare against GraphGA and MOLLEO, both previously evaluated on the same three docking targets under a 1,000-call oracle budget~\citep{wang2025efficient}.
All methods are evaluated under the same oracle budgets during optimization.
Oracle evaluations of candidates that fail the MMP reachability criterion also count toward the budget for every method.
Further details on the baseline methods are provided in Appendix~\ref{app:baseline_details}.

\textbf{Evaluation Metrics.}
For both tasks, a generated molecule is considered reachable if it forms an MMP edge with an initial molecule or a previously accepted molecule. Accepted molecules are added to the graph after each generation batch.
For property optimization, we report AUC@10 and AUC@100 under this criterion, together with the isolated node ratio (Iso).
Iso is the fraction of valid, scored unique generated molecules that do not connect to the initial graph under this reachability criterion.
AUC@$k$ measures the average running top-$k$ property score over the oracle budget, with higher values indicating better optimization performance.
For docking-based optimization, we report the same three metrics, computing AUC from docking scores capped at zero, $\bar{s}_{\mathrm{dock}}=\min(0,s_{\mathrm{dock}})$, where $s_{\mathrm{dock}}$ is the Vina docking score. Lower docking AUC and lower Iso indicate better performance.


\newcommand{\best}[1]{\textbf{#1}}
\newcommand{\second}[1]{\underline{#1}}

\subsection{Main Results}
\textbf{Task 1: Property Optimization.}
Table~\ref{tab:property_optimization_main} reports mean property optimization results; results with standard deviations are provided in Appendix~\ref{app:task1_std}.

\begin{table}[!htbp]
\centering
\fontsize{6.5}{8}\selectfont
\setlength{\tabcolsep}{1.15pt}
\renewcommand{\arraystretch}{1.05}
\caption{Property optimization on ZINC250 (10,000 oracle calls) and PMV21 (1,000 calls). Bold and underlined values denote the best and second-best results, respectively.}
\label{tab:property_optimization_main}
\begin{tabular*}{\textwidth}{@{\extracolsep{\fill}}cl
*{3}{c}!{\hspace{2pt}\vrule\hspace{2pt}}
*{3}{c}!{\hspace{2pt}\vrule\hspace{2pt}}
*{3}{c}!{\hspace{2pt}\vrule\hspace{2pt}}
*{3}{c}}
\toprule
 & Method & \multicolumn{3}{c!{\hspace{2pt}\vrule\hspace{2pt}}}{GSK3$\beta$} & \multicolumn{3}{c!{\hspace{2pt}\vrule\hspace{2pt}}}{QED} & \multicolumn{3}{c!{\hspace{2pt}\vrule\hspace{2pt}}}{JNK3} & \multicolumn{3}{c}{DRD2} \\
\cmidrule(lr){3-5}\cmidrule(lr){6-8}\cmidrule(lr){9-11}\cmidrule(lr){12-14}
 & & AUC$_{10}\uparrow$ & AUC$_{100}\uparrow$ & Iso$\downarrow$ & AUC$_{10}\uparrow$ & AUC$_{100}\uparrow$ & Iso$\downarrow$ & AUC$_{10}\uparrow$ & AUC$_{100}\uparrow$ & Iso$\downarrow$ & AUC$_{10}\uparrow$ & AUC$_{100}\uparrow$ & Iso$\downarrow$ \\
\midrule
\multirow{10}{*}{\rotatebox{90}{ZINC250}} & GraphGA & 0.3963 & 0.2589 & 0.8792 & 0.9304 & 0.8844 & 0.8092 & 0.5056 & 0.4081 & 0.8233 & 0.6163 & 0.3150 & 0.8772 \\
 & REINVENT & 0.6564 & 0.5338 & 0.6689 & 0.9392 & 0.9062 & 0.6274 & 0.7574 & 0.7097 & 0.3953 & 0.8903 & 0.8381 & 0.5380 \\
 & Aug-Mem & 0.6175 & 0.4400 & 0.9091 & 0.9366 & 0.8931 & 0.7300 & 0.7201 & 0.6445 & 0.7495 & 0.8552 & 0.7426 & 0.7895 \\
 & GP-BO & 0.8195 & 0.7198 & 0.6106 & 0.9231 & 0.8745 & 0.9330 & 0.3509 & 0.3050 & 0.3230 & 0.9083 & 0.8336 & \underline{0.3163} \\
 & MOLLEO & 0.8904 & \underline{0.8090} & 0.4642 & 0.9299 & 0.8751 & 0.7956 & \underline{0.7759} & \underline{0.7106} & 0.6214 & 0.7474 & 0.3679 & 0.8543 \\
 & NFBO & 0.8316 & 0.6026 & 0.9623 & 0.9384 & 0.8856 & 0.7273 & 0.6459 & 0.5559 & 0.6462 & 0.9179 & 0.7823 & 0.7489 \\
 & GenMol & 0.3360 & 0.1083 & 0.9901 & 0.9103 & 0.8266 & 0.8720 & 0.0003 & 0.0000 & 1.0000 & \underline{0.9860} & \underline{0.9522} & 0.4267 \\
 & RetMol & 0.5112 & 0.4687 & \underline{0.4019} & \underline{0.9457} & \underline{0.9396} & \underline{0.3427} & 0.5422 & 0.4656 & \underline{0.1996} & 0.9484 & 0.8536 & 0.4884 \\
 & InVirtuoGen & \underline{0.9012} & 0.7397 & 0.7102 & 0.9288 & 0.8646 & 0.8655 & 0.6833 & 0.5412 & 0.7918 & 0.6805 & 0.1792 & 0.7820 \\
 & MolWorld & \textbf{0.9576} & \textbf{0.9044} & \textbf{0.0460} & \textbf{0.9481} & \textbf{0.9470} & \textbf{0.0438} & \textbf{0.8225} & \textbf{0.7770} & \textbf{0.1105} & \textbf{0.9943} & \textbf{0.9819} & \textbf{0.0202} \\
\midrule
\multirow{10}{*}{\rotatebox{90}{PMV21}} & GraphGA & 0.2751 & 0.1856 & \underline{0.6057} & 0.7286 & 0.5408 & 0.8007 & 0.1490 & 0.0982 & 0.5527 & 0.7722 & 0.4403 & 0.6247 \\
 & REINVENT & None & None & 1.0000 & None & None & 1.0000 & None & None & 1.0000 & None & None & 1.0000 \\
 & Aug-Mem & None & None & 1.0000 & None & None & 1.0000 & None & None & 1.0000 & None & None & 1.0000 \\
 & GP-BO & 0.2426 & 0.1512 & 0.6597 & 0.6908 & 0.5429 & \underline{0.6237} & 0.1364 & 0.0858 & 0.6420 & 0.7674 & 0.4145 & \underline{0.5930} \\
 & MOLLEO & 0.3008 & \underline{0.1936} & 0.6603 & 0.7880 & \underline{0.5649} & 0.7903 & 0.1744 & \underline{0.1114} & \underline{0.5347} & 0.8019 & 0.4521 & 0.7047 \\
 & NFBO & 0.2733 & 0.1082 & 0.8463 & 0.6115 & 0.3218 & 0.8445 & 0.1662 & 0.0951 & 0.6734 & 0.6783 & 0.3092 & 0.7897 \\
 & GenMol & \underline{0.3138} & 0.0538 & 0.9810 & \underline{0.8010} & 0.3158 & 0.9477 & \underline{0.2315} & 0.0427 & 0.9717 & \underline{0.8659} & 0.3765 & 0.9123 \\
 & RetMol & 0.2037 & 0.0866 & 0.8917 & 0.6290 & 0.4175 & 0.9000 & 0.1122 & 0.0529 & 0.8930 & 0.7774 & \underline{0.5135} & 0.6157 \\
 & InVirtuoGen & 0.2287 & 0.1284 & 0.9000 & 0.6113 & 0.4287 & 0.8973 & 0.1321 & 0.0741 & 0.9000 & 0.4878 & 0.2112 & 0.7478 \\
 & MolWorld & \textbf{0.4225} & \textbf{0.3490} & \textbf{0.0493} & \textbf{0.8892} & \textbf{0.8194} & \textbf{0.0744} & \textbf{0.2710} & \textbf{0.2175} & \textbf{0.0437} & \textbf{0.9018} & \textbf{0.8205} & \textbf{0.0990} \\
\bottomrule
\end{tabular*}
\end{table}
Across both ZINC250 and PMV21, \textsc{MolWorld} achieves the strongest overall performance, obtaining the best or highly competitive AUC scores across most tasks. 
As de novo generation methods, REINVENT and Augmented Memory do not explicitly enforce connectivity to the initial graph.
Although some of their generated molecules satisfy the MMP connectivity constraint on ZINC250, none do on PMV21; their corresponding AUC entries are therefore reported as None.
\textsc{MolWorld} achieves the lowest isolated node ratio across all eight dataset--property combinations, indicating that the generated molecules remain well integrated into the molecule-transfer graph through valid local transformations.

\textbf{Task 2: Docking-Based Optimization.}
Table~\ref{tab:docking_optimization_main} reports docking-based optimization results on A2AR, DRD3, and EGFR. \textsc{MolWorld} achieves the lowest AUC@10 and AUC@100 across all three targets among the compared methods, demonstrating effective discovery of MMP-connected molecules with favorable docking scores over the search budget. It also consistently achieves the lowest isolated node ratio, indicating that its generated molecules are better integrated into the molecule-transfer graph. Together, these results show that \textsc{MolWorld} combines effective docking optimization with structurally connected molecular generation.

\begin{table}[!htbp]
\centering
\scriptsize
\setlength{\tabcolsep}{4pt}
\renewcommand{\arraystretch}{1.08}
\caption{Docking optimization with 1,000 oracle calls (lower is better).}
\label{tab:docking_optimization_main}
\resizebox{\textwidth}{!}{%
\begin{tabular}{l*{9}{c}}
\toprule
Method & \multicolumn{3}{c}{A2AR} & \multicolumn{3}{c}{DRD3} & \multicolumn{3}{c}{EGFR} \\
\cmidrule(lr){2-4}\cmidrule(lr){5-7}\cmidrule(lr){8-10}
 & AUC@10$\downarrow$ & AUC@100$\downarrow$ & Iso$\downarrow$ & AUC@10$\downarrow$ & AUC@100$\downarrow$ & Iso$\downarrow$ & AUC@10$\downarrow$ & AUC@100$\downarrow$ & Iso$\downarrow$ \\
\midrule
GraphGA & -9.473 & -8.237 & \underline{0.765} & -8.644 & -7.515 & 0.828 & -9.069 & -7.794 & \underline{0.799} \\
MOLLEO & \underline{-9.952} & \underline{-8.296} & 0.779 & \underline{-9.055} & \underline{-7.663} & \underline{0.807} & \underline{-9.209} & \underline{-7.826} & 0.823 \\
\textsc{MolWorld} & \textbf{-10.351} & \textbf{-8.908} & \textbf{0.121} & \textbf{-9.687} & \textbf{-8.646} & \textbf{0.119} & \textbf{-9.502} & \textbf{-8.330} & \textbf{0.159} \\
\bottomrule
\end{tabular}%
}
\end{table}

\subsection{Understanding \textsc{MolWorld}}

\begin{figure}[!htbp]
  \centering
  \includegraphics[width=\textwidth]{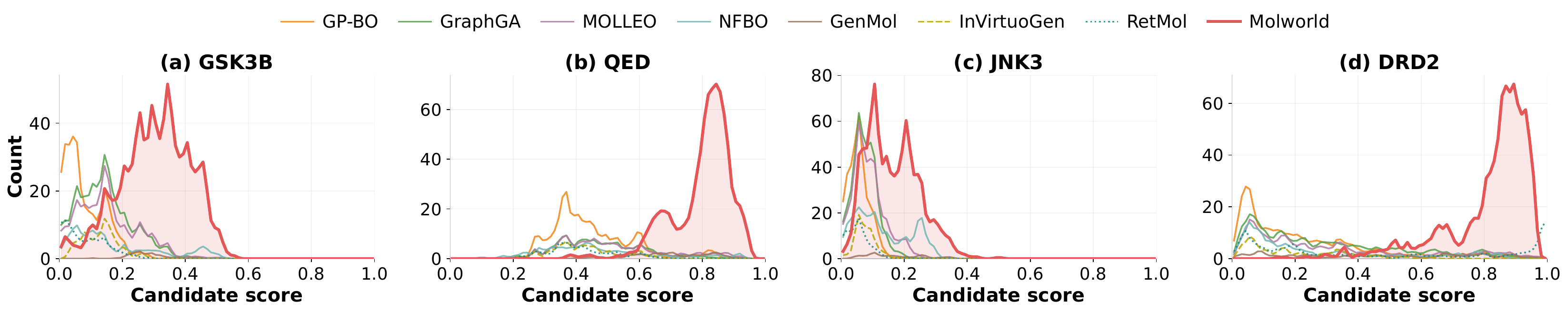}
  \caption{Property score distributions of MMP-connected molecules on PMV21. Curves show smoothed molecule counts across score values under a 1,000-call oracle budget.}
  \label{fig:dis}
\end{figure}

\textbf{Score Distribution and Optimization Dynamics.}
To further analyze the quality of generated molecules, we visualize the score distributions of reachable molecules under a fixed oracle budget on PMV21 (Figure~\ref{fig:dis}).
The plotted distributions show that \textsc{MolWorld} generates more molecules in high-score regions, particularly on QED and DRD2.
This illustrative view complements top-$k$ AUC metrics by showing molecule counts across the score range. We further analyze the optimization process using the top-10 running average score as a function of oracle calls (Appendix~\ref{sec:convergence}).
\textsc{MolWorld} converges faster on several tasks, particularly GSK3$\beta$ and JNK3 on ZINC, while remaining comparable to strong baselines on the remaining tasks.

\begin{figure}[!htbp]
  \centering
  \includegraphics[width=\textwidth]{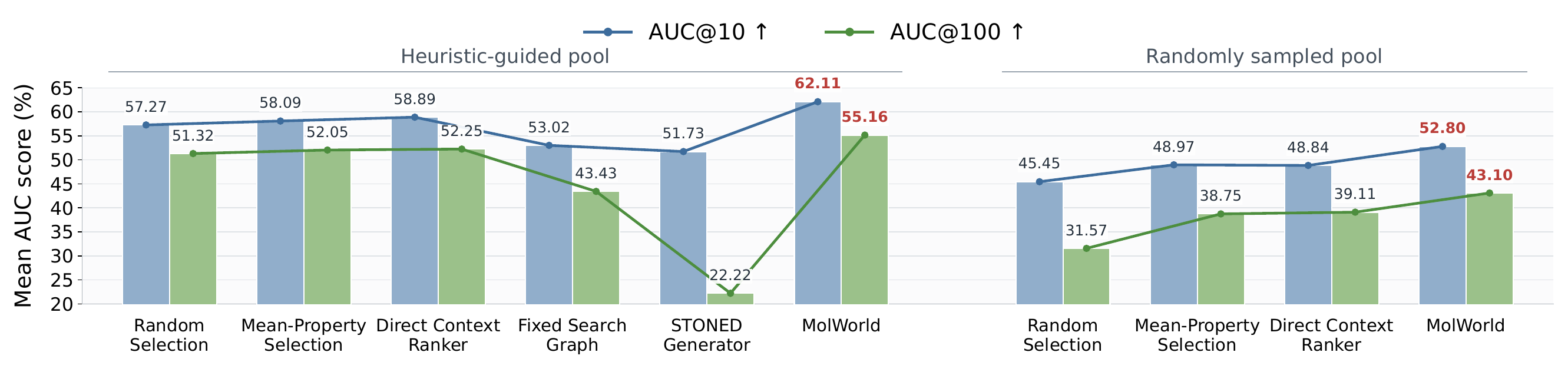}
  \caption{Property optimization ablations on PMV21, averaged across GSK3$\beta$, QED, JNK3, and DRD2, with 1,000 oracle evaluations per task.}
  \label{fig:property_ablation}
\end{figure}

\textbf{Ablation Study.}
We examine the contributions of molecule world model-guided anchor context selection, context-conditioned generation, and iterative graph update through five variants (Figure~\ref{fig:property_ablation}). \textbf{Random Anchor Selection} randomly selects contexts from the same candidate pool, while \textbf{Mean-Property Selection} ranks them by the mean observed property value of their constituent molecules. These variants test whether the candidate construction procedure and existing property observations are sufficient for effective selection. \textbf{Direct Context Ranker} replaces the molecule world model and its property scorer with a learned ranker that directly takes the current graph and anchor representations as input. It uses the same supervision and data split, with a comparable trainable head size, testing whether directly learning to rank contexts is sufficient without a molecule world model. These selection variants share the context-conditioned generator. \textbf{Fixed Search Graph} retains the learned selection and generation modules but prevents newly generated molecules from becoming part of subsequent generation contexts. \textbf{STONED Generator} replaces the context-conditioned generator with random SELFIES mutation of a molecule sampled from the selected context. To examine the influence of candidate-pool construction, we additionally evaluate \textsc{MolWorld}, Random Anchor Selection, Mean-Property Selection, and Direct Context Ranker using a pool of randomly sampled connected five-molecule contexts. This replaces the original heuristic-guided pool construction while retaining each method's context-selection strategy. All variants use the same setup.

As shown in Figure~\ref{fig:property_ablation}, \textsc{MolWorld} achieves the highest average AUC@10 and AUC@100 across the four properties. Its improvement over Direct Context Ranker supports predicting structural expansion before assessing property quality, rather than ranking contexts directly. The performance drop with Fixed Search Graph highlights the importance of reusing newly accepted molecules as anchors for further exploration. Replacing our generator with STONED reduces average AUC@100 from 0.5516 to 0.2222, supporting the contribution of the learned context-conditioned generator to optimization performance. \textsc{MolWorld} also achieves the highest average AUC@10 and AUC@100 among the compared selection strategies under randomly sampled candidate pools, indicating that its selection advantage persists when heuristic pool construction is removed.

\begin{figure}[!htbp]
  \centering                 

  \includegraphics[width=1\textwidth]{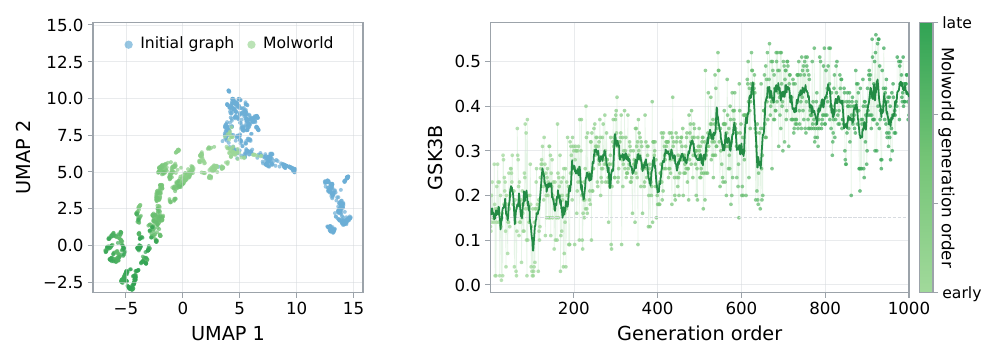} 
  \caption{Generated molecule distribution and oracle-score progression on GSK3$\beta$.}

  \label{fig:gsk}          
\end{figure}

\textbf{Generated Molecule Distribution.}
Figure~\ref{fig:gsk} shows a UMAP projection of initial and generated molecules (left) and oracle scores over generation order (right) on GSK3$\beta$.

In the UMAP projection, early generated molecules largely overlap with the initial set, while later molecules gradually expand into new regions of chemical space.
The 10-step moving-average curve shows improving oracle scores over generation order.
Appendix~\ref{app:graph_extension} shows similar patterns on QED, JNK3, and DRD2.
Anchor contexts quickly shift from initial to newly generated molecules after the early iterations, enabling \textsc{MolWorld} to reuse its discoveries for further graph expansion toward higher-scoring regions (Appendix~\ref{sec:anchor_evolution}).

\begin{figure}[!htbp]
  \centering
  \includegraphics[width=0.9\textwidth]{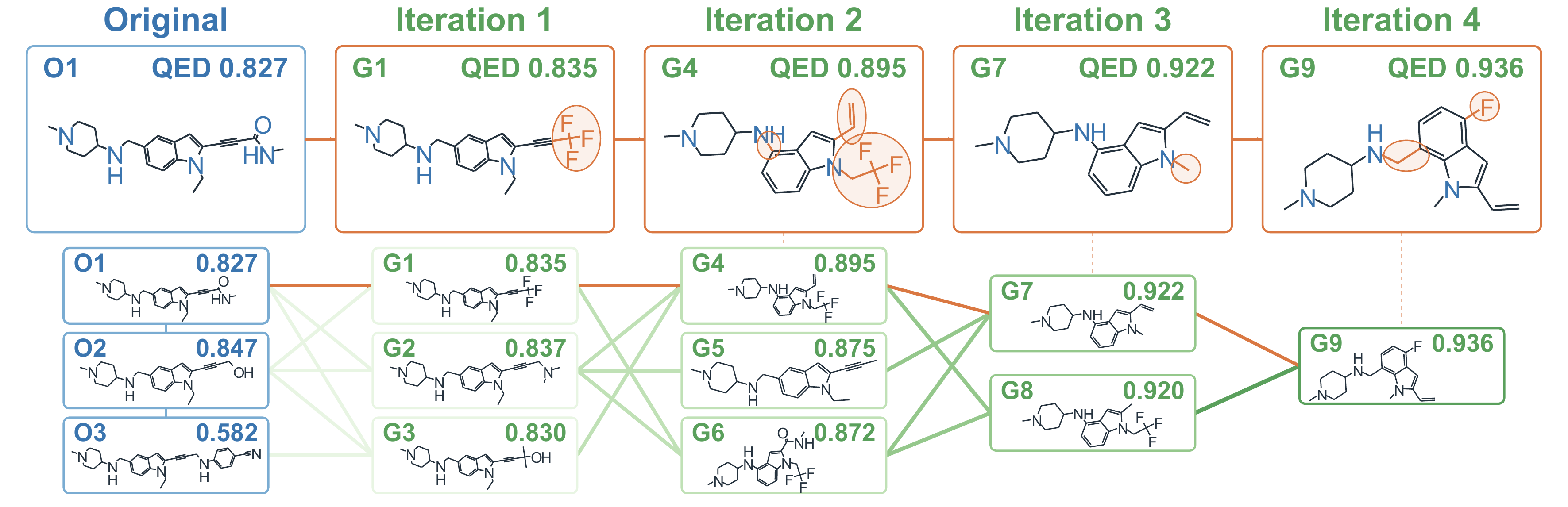}
  \caption{Sequential QED optimization with \textsc{MolWorld}: the molecular graph (bottom) and a selected sequence (top). Orange circles highlight selected structural differences.}
  \label{fig:vis}
  \label{fig:combined}
\end{figure}

\textbf{Qualitative Visualization of Sequential Molecular Optimization.}
Figure~\ref{fig:vis} presents initial molecules ($O1$--$O3$) and generated candidates across four iterations of QED optimization.
The upper panels compare the selected sequence $O1$, $G1$, $G4$, $G7$, and $G9$, with QED scores of $0.827$, $0.835$, $0.895$, $0.922$, and $0.936$, respectively.
The selected molecules retain an indole core and an $N$-methylpiperidine moiety, while differing in their terminal groups, $N$-substituents, and side-chain attachment patterns.
For the depicted structures, the descriptor contributions in the RDKit QED calculation help explain the score progression within this composite objective \citep{bickerton2012quantifying}.
Relative to $O1$, $G1$ replaces the terminal amide group with a trifluoromethyl group.
The hydrogen-bond donor count decreases from two to one, improving its QED desirability term; less favorable changes in other terms partly offset this gain, yielding a modest net increase.
$G4$ differs from $G1$ at several positions, including the substituent at the indole nitrogen and the amine-side-chain attachment.
The absence of the alkyne motif removes a match to the structural-alert patterns used by RDKit QED, providing the largest positive contribution to this score increase.
In $G7$, the indole $N$-substituent is a methyl group rather than the trifluoroethyl group in $G4$.
The calculated logP decreases from $4.35$ to $3.33$, and the corresponding improvement in the QED lipophilicity term is the main contributor to the higher score.
$G9$ additionally differs from $G7$ in aryl fluorination and the attachment position and linker length of the aminoalkyl side chain.
Its further QED increase is driven primarily by a more favorable molecular-weight term, with a smaller favorable contribution from calculated logP.
The score increases reflect the QED objective and do not, by themselves, establish improvements in biological activity or experimental developability.

\FloatBarrier
\section{Conclusion}
We introduced \textsc{MolWorld}, a molecule world model-guided framework that optimizes molecular properties by expanding a molecule-transfer graph while preserving MMP reachability from the initial molecular set. Its molecule world model predicts graph expansion from candidate anchor contexts in latent space, with structural dynamics shared across properties. A property-specific scorer evaluates these predictions to guide anchor context selection. The context-conditioned generator uses molecular structures and MMP relations in the selected contexts to propose molecules, which enter the graph through verified MMP connections. Experiments on property optimization and docking-based optimization show that \textsc{MolWorld} discovers high-scoring molecules while maintaining stronger connectivity than competing methods. Ablation studies and further analyses support the effectiveness of its components.


\subsection*{AI use statement}

We used generative AI tools for manuscript editing and coding assistance, including debugging and scripts for processing and visualizing experimental results.

\subsection*{Ethics statement}

This research aims to support drug discovery through computational molecular optimization. The experiments evaluate molecular candidates using computational property predictors and docking oracles. These evaluations support candidate prioritization for further investigation, while practical applications require appropriate experimental validation and safety assessment.

\subsection*{Reproducibility statement}

Section~\ref{sec:method} describes the method. Appendix~\ref{app:implementation_details} details the structural transition dataset construction and splits, model architectures, training objectives, hyperparameters, and anchor context selection. The Experimental Setup and Additional Experimental Details sections specify evaluation datasets, oracle budgets, the reachability criterion, metrics, and the docking protocol.

\bibliography{iclr2027_conference}
\bibliographystyle{iclr2027_conference}
\appendix

\section{Additional Related Work}\label{app:additional_related_work}

\textbf{World Models.} World models aim to build compact and actionable abstractions of the external world, enabling agents to reason about hidden dynamics, predict future states, and make decisions under uncertainty \cite{ding2025understanding}. Existing studies can be broadly divided into two lines. The first line treats world models as latent representations for decision-making, especially in model-based reinforcement learning. Early neural world models learn compressed visual states and recurrent dynamics to support policy learning in imagination~\citep{ha2018worldmodels}, while subsequent methods formulate control as learning latent dynamics and optimizing policies through imagined rollouts~\citep{hafner2020dreamer,hafner2023dreamerv3,hansen2024tdmpc2}. With the rise of large language and multimodal models, this view has been extended to embedding world knowledge in foundation models~\citep{gurnee2024spaceandtime}. 
The second line views world models as future simulators. Recent video generation models show that large-scale generative models can capture temporal coherence, object persistence, and some physical regularities in visual sequences~\citep{brooks2024sora}. Recent work also moved to interactive simulators and embodied environments, where models generate future states conditioned on actions, language instructions, or multimodal observations~\citep{yang2023unisim,bruce2024genie,hu2023gaia}, especially in autonomous driving and robotics, where world models serve both as implicit scene-understanding modules and as simulators for counterfactual prediction, planning, and policy learning~\citep{wang2024drivedreamer,lin2024dynalang}.
Incorporating world models into scientific discovery is a promising yet under-explored area. MDGen models full molecular trajectories as a time series of 3D structures and supports forward simulation~\citep{jing2024mdgen}. Recent work further formulates protein-folding pathway prediction as a generative world-modeling problem~\citep{ianeselli2025generative}. This direction suggests that the scope of world models can extend from visual and embodied prediction to scientific simulation, where learning controllable future dynamics may accelerate biomedical discovery.

\section{Implementation Details}
\label{app:implementation_details}

We provide implementation details for the structural transition dataset, molecule world model training, anchor context selection, property scoring, and generator training. Molecules are represented as nodes in a molecule-transfer graph, with MMP relations defining its edges. Offline collection uses reference-graph edges and RDKit MMP checks; online expansion uses RDKit MMP checks.

\subsection{Structural Transition Dataset}
\label{app:transition_dataset}

\textbf{Overview.} We construct an offline dataset of molecular graph expansions from the ChEMBL molecule-transfer graph. The dataset supports graph-encoder pretraining, action-conditioned dynamics learning, and subsequent property-scorer training. Each state-action pair records the current graph $G_t$, an anchor context $Z$, and one or more observed generation outcomes. An outcome includes the raw candidates $\mathcal{X}_{t,Z}$, accepted new molecules $\widetilde{\mathcal{X}}_{t,Z}$, verified MMP edges, and the resulting graph $G_{t+1}$. Structural data collection is independent of the target property: property labels are attached after the graph-expansion trajectories have been collected.

\textbf{Initial graphs and trajectories.} We sample initial graphs from ChEMBL at four scales to expose the encoder and dynamics model to different graph sizes. The configured collection contains 150 root trajectories:
\begin{center}
\begin{tabular}{rrr}
\toprule
Initial graph size & Depth-ten trajectories & Depth-twenty trajectories \\
\midrule
1,000 nodes & 48 & 12 \\
10,000 nodes & 24 & 6 \\
100,000 nodes & 36 & 9 \\
250,000 nodes & 12 & 3 \\
\bottomrule
\end{tabular}
\end{center}
Each root defines an initial graph and a branching expansion trajectory. At every depth, a bounded frontier retains at most four states for further expansion. The retained states cover structurally productive, intermediate, randomly exploratory, and structurally diverse outcomes. This provides varied transitions without exhaustively expanding the trajectory tree.

\textbf{Action sampling for data collection.} At each active state, we construct a pool of 1,000 unique connected five-molecule anchor contexts. Seed proposals alternate between uniform sampling, low-visit sampling, and low-, medium-, and high-degree sampling. Each seed is expanded through existing graph edges until five distinct connected nodes are obtained. The generator input uses a breadth-first spanning tree of the selected nodes, with four undirected edges supplied bidirectionally.

We score training-data candidates using structural information:
\begin{equation}
\label{eq:app_data_structural_score}
b_{\mathrm{struct}}(Z)
=\frac{1}{|V_Z|}\sum_{v\in V_Z}\log(1+\deg(v))
+\frac{0.35}{1+\overline c_Z}
+0.15\,\frac{\sigma_{\deg,Z}}{1+\overline{\deg}_Z},
\end{equation}
where $\overline c_Z$ is the mean visit count, and $\overline{\deg}_Z$ and $\sigma_{\deg,Z}$ are the mean and standard deviation of node degrees within the context.

The pool is divided into low-, medium-, and high-score groups. We select three actions from the high group, two from the medium group, two from the low group, one random action, and two additional actions favoring diversity. Within the structural groups, selection combines normalized structural scores with diversity relative to already selected contexts. Diversity combines node-set Jaccard distance and a difference in mean degree, with weights $0.8$ and $0.2$, respectively.

This procedure collects ten actions per state without target-property information, unlike downstream property-guided candidate construction (Appendix~\ref{app:anchor_selection}).

\textbf{Generation and structural acceptance.} The pretrained context-conditioned generator samples 16 sequences per action using nucleus sampling with $p=0.95$, temperature $1.0$, and maximum sequence length 64. Twenty percent of state-action pairs are executed with three random seeds; the remaining pairs use one seed. All seed outcomes of the same action share a state-action identifier.

Generated strings are canonicalized, checked for molecular validity, and deduplicated. Molecules already present in the current graph are excluded from the accepted-new-molecule set. For generated molecules belonging to the ChEMBL reference graph, insertion relations are retrieved from the reference edges. Outside that graph, MMP relations are computed with RDKit \texttt{rdMMPA}, using at most three cuts and 20 cut bonds, and a variable-fragment heavy-atom ratio capped at $0.33$.

We apply the same connectivity-based acceptance rule used during online graph expansion (Appendix~\ref{app:graph_update}). A valid new molecule is accepted if it forms a verified MMP edge with a molecule in the current graph. Accepted molecules are then incorporated to obtain the next graph state. Verified edges linking accepted new molecules to existing or other accepted molecules are retained. No learned link predictor participates in data collection. The stored records retain the structural evidence associated with added edges.

For actions with multiple seeds, one outcome is randomly chosen as the action's continuation representative. The other outcomes remain available for learning the distribution of results associated with that state-action pair.

\textbf{Data splits and training cache.} Root trajectories are assigned to training, validation, and test splits using a deterministic hash-based rule with nominal proportions $80/10/10$. All states, actions, and seed outcomes descended from a root inherit its split. Thus, transitions within one root trajectory cannot appear in different splits. The split is trajectory-based; it does not enforce molecule-disjoint or scaffold-disjoint partitions between independently sampled roots.

The cache used for dynamics and property-scorer training contains 124 training roots and 11 validation roots, totaling 6,035 action groups, 60,350 state-action pairs, and 84,490 action-seed outcomes. Test roots are excluded from this cache and from checkpoint selection. Initial graphs provide graph-encoder pretraining examples, while the recorded state-action outcomes provide dynamics supervision. Property scores of accepted new molecules provide scorer labels.

\subsection{Molecule World Model Training}
\label{app:world_model_details}

\textbf{Training stages.} We first pretrain a graph encoder using masked latent prediction on the initial graphs of the structural dataset. Its target encoder is then frozen and reused as $E_G$ for both current graphs and local expansion outcomes. The anchor encoder $E_A$ is extracted from the pretrained context-conditioned generator and remains frozen. We train the dynamics model $\mathcal{W}_\phi$ on recorded expansions, then freeze it when training the property scorer.

\subsubsection{Graph Encoder}

\textbf{Molecular features and message passing.} Each molecule is represented by a radius-two, 2,048-bit Morgan fingerprint and 23 RDKit descriptors. Active fingerprint bits are embedded and averaged to obtain a 192-dimensional fingerprint representation. A separate MLP maps the descriptors to 128 dimensions. Their concatenation is projected to a 256-dimensional molecular representation.

The descriptors are molecular weight, LogP, molar refractivity, TPSA, Labute ASA, heavy-atom count, heteroatom count, hydrogen-bond donor and acceptor counts, rotatable-bond count, total ring count, aromatic, aliphatic, and saturated ring counts, fraction of $sp^3$ carbons, NHOH count, NO count, valence-electron count, Bertz complexity, Hall--Kier alpha, and the three Kappa indices. Descriptor preprocessing uses the fixed statistics stored with the reusable molecular-feature cache.

Two sparse message-passing layers incorporate molecule-transfer graph connectivity. Each layer combines a transformed self-representation with a transformed mean-neighbor representation, followed by a SiLU activation, dropout, a residual connection, and layer normalization.

\textbf{Graph tokens.} The encoder summarizes a graph with 16 tokens of dimension 256. Four tokens contain the mean, standard deviation, elementwise maximum, and degree-weighted mean of the node representations. The degree-weighted summary uses weights proportional to $\log(1+\deg(v))$.

Twelve additional tokens are obtained by learned competitive attention pooling. Each node is assigned across the learned queries using a softmax with temperature $0.25$; assignments are then normalized over nodes for each pooled token. The resulting tokens receive layer normalization. This combines global statistics with learned summaries of different regions of the graph.

\textbf{Masked latent prediction.} Following teacher--student latent prediction methods~\cite{baevski2022data2vec,assran2023self}, we maintain an online encoder $E_\eta$ and an exponential-moving-average target encoder $E_{\bar\eta}$. The online encoder receives a graph with the molecular content of 15\% of nodes masked by a learned token, retaining edges. The target encoder receives the unmasked graph.

Let $M$ denote the masked node set, $u_v$ and $H$ the online node and graph representations, and $\bar u_v$ and $\bar H$ their target counterparts. Separate heads $P_n$ and $P_g$ predict the node and graph representations. To discourage redundant graph tokens, we also penalize similarity between normalized predicted graph tokens $q_i$, with $K=16$. The prediction losses, diversity penalty, and combined objective are
\begin{equation}
\label{eq:app_encoder_loss}
\begin{aligned}
\mathcal{L}_{\mathrm{node}}
&=\frac{1}{|M|}\sum_{v\in M}
\ell_{\mathrm H}\!\left(P_n(u_v),\operatorname{sg}[\operatorname{LN}(\bar u_v)]\right),\\
\mathcal{L}_{\mathrm{graph}}
&=\ell_{\mathrm H}\!\left(P_g(H),\operatorname{sg}[\operatorname{LN}(\bar H)]\right),\\
\mathcal{L}_{\mathrm{div}}
&=\frac{1}{K(K-1)}\sum_{i\ne j}(q_i^\top q_j)^2,\\
\mathcal{L}_{\mathrm{enc}}
&=\mathcal{L}_{\mathrm{node}}+0.25\,\mathcal{L}_{\mathrm{graph}}+0.05\,\mathcal{L}_{\mathrm{div}}.
\end{aligned}
\end{equation}
Here, $\operatorname{sg}$ denotes stop-gradient and $\operatorname{LN}$ denotes layer normalization. The smooth-$L_1$ loss averages the following penalty over representation coordinates:
\begin{equation}
\label{eq:app_smooth_l1}
\ell_{\mathrm H}(x,y)=\operatorname{mean}_k
\begin{cases}
\frac12(x_k-y_k)^2,& |x_k-y_k|<1,\\
|x_k-y_k|-\frac12,& \text{otherwise}.
\end{cases}
\end{equation}
Only the online encoder and prediction heads receive gradient updates. The target encoder follows the exponential-moving-average update $\bar\eta\leftarrow m\bar\eta+(1-m)\eta$, with momentum increasing from $0.9995$ toward $1.0$ during training.

\textbf{Optimization and initialization.} The final competitive-pooling encoder is initialized from an earlier graph JEPA checkpoint by transferring compatible parameters. The earlier checkpoint uses the same molecular encoder and two graph layers and is trained with node and graph prediction losses. The competitive-pooling stage adds multi-token pooling and token-diversity regularization.

Both stages use AdamW with learning rate $2\times10^{-4}$, weight decay $0.04$, 20 epochs, and two warmup epochs followed by cosine learning-rate decay. Training balances the sampling of initial graphs across graph-size groups. Checkpoint selection uses the combined validation objective. The selected target encoder from the competitive-pooling stage is frozen and used as $E_G$.

\textbf{Validation.} We evaluate whether the trained encoder uses MMP relations by removing or shuffling the input edges on the validation graphs. We keep the model parameters, masked nodes, and target representations fixed across settings; the target encoder always receives the unmasked graph with its original edges. We report the masked-node prediction loss $\mathcal{L}_{\mathrm{node}}$ and graph representation prediction loss $\mathcal{L}_{\mathrm{graph}}$, averaged over validation graphs. Both measure smooth-$L_1$ error against the target encoder's representations, with lower values indicating more accurate predictions.

\begin{table}[htbp]
\centering
\caption{Graph encoder validation under input-edge perturbations.}
\label{tab:encoder_edge_validation}
\small
\begin{tabular}{lcc}
\toprule
Input edges & Node prediction loss $\downarrow$ & Graph prediction loss $\downarrow$ \\
\midrule
Original MMP edges & \textbf{0.1048} & \textbf{0.0365} \\
Removed edges & 0.3617 & 0.1269 \\
Shuffled edges & 0.3382 & 0.0429 \\
\bottomrule
\end{tabular}
\end{table}

Removing or shuffling the edges increases masked-node prediction error to 3.45 and 3.23 times that obtained with the original edges, respectively (Table~\ref{tab:encoder_edge_validation}). This indicates that the encoder uses the observed MMP relationships to predict masked molecular representations.

\subsubsection{Anchor Encoder}

The anchor representation is extracted from the same molecular encoding and graph-fusion modules used by the context-conditioned generator. Molecular token representations receive graph-conditioned updates through the generator's fusion operation. Averaging over valid sequence tokens separately for each anchor molecule gives $a_Z=E_A(Z)\in\mathbb{R}^{5\times768}$, with one action token per molecule preserving the generator's learned contextual representation. The encoder is not retrained for dynamics learning. See Section~\ref{sec:generator} and Appendix~\ref{app:generator_details} for generator architecture and pretraining.

\subsubsection{Dynamics Model}

\textbf{Local expansion target.} For state-action pair $(G_t,Z)$ and generation seed $r$, let $\Delta G_{t,Z}^{(r)}$ denote the recorded local outcome graph. Its nodes consist of all accepted new molecules and all existing molecules incident to the saved insertion edges. Its edges contain every saved new-to-old and new-to-new relation. Co-occurrence in this local graph does not add old-to-old edges.

The target and prediction are
\begin{equation}
\delta_Z=\frac{1}{R_Z}\sum_{r=1}^{R_Z}E_G\!\left(\Delta G_{t,Z}^{(r)}\right),
\qquad
\widehat\delta_Z=\mathcal{W}_\phi\!\left(E_G(G_t),E_A(Z)\right),
\label{eq:app_dynamics_target}
\end{equation}
where $R_Z\in\{1,3\}$. An outcome with no accepted molecules receives an all-zero token representation. The target is an encoding of the local expansion, rather than the difference between current-graph and next-graph embeddings.

The complete current graph and local outcome graph use exactly the same frozen encoder. New-node identity is used to construct the local outcome graph but is not supplied as an additional encoder feature. For each distinct current graph, message passing and graph pooling are recomputed; only graph-independent molecular features are cached.

\textbf{Architecture.} The dynamics model receives 16 current-graph tokens of dimension 256 and five anchor tokens of dimension 768. A learned projection maps the anchor tokens to dimension 256. These inputs are concatenated with 16 learned output queries, with positional and type embeddings distinguishing the three token groups.

A four-layer Transformer with eight attention heads and dropout $0.1$ processes the sequence. An output MLP and layer normalization transform the output-query representations into the predicted expansion tokens $\widehat\delta_Z\in\mathbb{R}^{16\times256}$.

\textbf{Training.} We minimize
\begin{equation}
\mathcal{L}_{\mathrm{dyn}}=\frac{1}{|\mathcal{D}|}
\sum_{(G_t,Z)\in\mathcal{D}}\ell_{\mathrm H}(\widehat\delta_Z,\delta_Z),
\label{eq:app_dynamics_loss}
\end{equation}
where $\mathcal{D}$ contains training state-action pairs. Both encoders remain frozen.

The base configuration uses AdamW with learning rate $3\times10^{-4}$, weight decay $0.04$, 30 epochs, and eight state groups per batch. Training uses a 5\% warmup followed by cosine decay, gradient clipping at norm 1, and mixed precision. Checkpoint selection minimizes validation dynamics loss.

\textbf{Validation.} We evaluate the selected dynamics checkpoint by comparing its predicted expansion representations $\widehat{\delta}_Z$ with the observed expansion targets $\delta_Z$ on the validation set. Prediction error is measured by smooth-$L_1$ loss, averaged over representation coordinates, tokens, and state--context pairs. As a constant baseline, we predict the mean expansion target computed from the training set. We also evaluate the trained model with mismatched contexts by cyclically shifting anchor contexts within each validation batch while keeping graph inputs and targets unchanged.

\begin{table}[htbp]
\centering
\caption{Validation of latent graph expansion predictions.}
\label{tab:dynamics_prediction_validation}
\small
\begin{tabular}{lc}
\toprule
Prediction setting & Smooth-$L_1$ loss $\downarrow$ \\
\midrule
Mean training-set target & 0.2083 \\
Dynamics model with mismatched contexts & 0.2148 \\
Dynamics model with matched contexts & \textbf{0.0488} \\
\bottomrule
\end{tabular}
\end{table}

With matched contexts, the dynamics model reduces prediction error by 76.57\% relative to the mean-target baseline (Table~\ref{tab:dynamics_prediction_validation}). Its mean cosine similarity is 0.9421, computed between corresponding predicted and target tokens and averaged over tokens and validation pairs. Higher error under context mismatch indicates that accurate expansion prediction depends on the anchor context.

\subsection{Anchor Context Selection and Property Scoring}
\label{app:anchor_selection}

\subsubsection{Candidate Construction During Optimization}

The online candidate pool uses current property observations to focus search, while visit-aware random proposals and repeated-use penalties promote exploration. This differs from the property-independent action sampling used to collect the structural training dataset.

\textbf{Search region and seeds.} Only connected components containing at least five nodes are eligible. When more than 3,000 nodes are eligible, we attempt to restrict search to the subgraph induced by the 3,000 highest-property nodes. If no connected component in this subgraph can support a five-node anchor, we retain the original eligible search region.

Seed selection combines up to 100 highest-property nodes with up to 200 additional nodes sampled preferentially from less-visited candidates. To sample $n$ such nodes, candidates are sorted by cumulative visit count with randomized tie-breaking. The first $5n$ candidates, or all available candidates if fewer, form a shortlist that is shuffled before selecting $n$ nodes. Previously selected seeds are excluded from this additional sample.

\textbf{Connected-subgraph beam search.} Each partial context is expanded by adding an adjacent node. The expansion candidates consist of up to ten highest-property frontier nodes and ten additional frontier nodes selected through the low-visit sampling procedure. Equivalent node sets are deduplicated.

Let $c_t(v)$ be the number of previously selected anchors containing $v$, and let $r_t(v)$ count its occurrences in selected anchors from the previous five iterations. For cached property values $f(v)$, define the context mean $\overline f_Z=|V_Z|^{-1}\sum_{v\in V_Z}f(v)$. Partial contexts are scored by
\begin{equation}
\label{eq:app_candidate_beam_score}
b_t(Z)=\overline f_Z-\lambda_v P_v(Z)-\lambda_o P_o(Z)-\lambda_r P_r(Z).
\end{equation}
Writing $R_t(Z)=\{v\in V_Z:r_t(v)>0\}$ for recently used nodes, the visit, overlap, and repeated-use penalties are
\begin{equation}
\label{eq:app_candidate_penalties}
\begin{aligned}
P_v(Z)&=\frac{1}{|V_Z|}\sum_{v\in V_Z}\log(1+c_t(v)),\\
P_o(Z)&=\max\!\left(0,\,|R_t(Z)|-3\right),\\
P_r(Z)&=\frac{1}{\max\{1,|R_t(Z)|\}}\sum_{v\in R_t(Z)}\log(1+r_t(v)).
\end{aligned}
\end{equation}
The empty sum gives $P_r(Z)=0$ when no context node was used recently. The default coefficients are $\lambda_v=0.02$, $\lambda_o=0.03$, and $\lambda_r=0.01$. At each depth, the beam retains at most 300 partial contexts. Search stops growing a context once it contains five nodes. Complete anchors already executed in earlier iterations are excluded from the candidate pool.

\textbf{Adaptive penalties.} The implementation adjusts penalties according to recorded generation yield. For an iteration with nonempty output but fewer than ten newly recorded candidates, each coefficient is multiplied by $1.25$, up to respective maxima of $0.08$, $0.12$, and $0.04$. For nonempty output of at least ten candidates, coefficients are multiplied by $0.95$, bounded below by their defaults. Empty-output iterations do not update these coefficients.

\textbf{Final selection.} Every completed candidate is scored by the property scorer using its current-graph and predicted-expansion representations. Final selection takes the highest-scoring $B$ candidates without additional MMR, property-based, or repeated-use reranking. Thus, heuristic scores determine candidate availability, while the learned scorer determines selection within the resulting pool.

The runner defaults to $B=20$. If no explicit batch size is supplied, an adaptive setting can increase it to 50 when the preceding five iterations produce fewer than 200 recorded candidates. Explicit batch-size settings override this adaptation. Candidate-pool and batch sizes can also be set per run.

\subsubsection{Property Scorer}

\textbf{Training labels.} For generation seed $r$, let $\widetilde{\mathcal{X}}_{t,Z}^{(r)}$ denote the accepted new molecules for action $Z$, and let $f_{(k)}^{(r)}$ be their $k$-th highest property score. Missing entries are assigned zero. We define the seed-level label and aggregated action label as
\begin{equation}
y_t^{(r)}(Z)=\frac{1}{3}\sum_{k=1}^{3}f_{(k)}^{(r)},
\qquad
y_t(Z)=\frac{1}{R_Z}\sum_{r=1}^{R_Z}y_t^{(r)}(Z).
\label{eq:app_property_label}
\end{equation}
For an action producing no accepted molecules, the label is zero. Keeping the denominator fixed at three makes the label sensitive to both property quality and accepted-molecule yield. The averaging is performed after computing each seed's label, rather than pooling molecules across seeds. Labels use only newly accepted molecules; properties of molecules already present in $G_t$ do not contribute. Property-specific scorers share the structural encoder and dynamics model.

\textbf{Architecture.} The property scorer receives 16 current-graph tokens and 16 predicted-expansion tokens, each of dimension 256. These are concatenated with one learned query token. Type and positional embeddings distinguish the current graph, predicted expansion, and query. A three-layer Transformer with eight attention heads and dropout $0.1$ processes the sequence, and an MLP maps the query output to the scalar action score $s_t(Z)=Q_\psi(h_t,\widehat\delta_Z)$. Scorer training uses dynamics predictions $\widehat\delta_Z$, not the observed outcome targets $\delta_Z$. The raw anchor representation is not provided as a separate scorer input. The graph encoder, anchor encoder, and dynamics model remain frozen.

\textbf{Ranking objectives.} Losses compare actions belonging to the same current state. For a group with labels $y_i$ and predicted scores $s_i$, define the action probabilities $p_i=\exp(s_i)/\sum_j\exp(s_j)$, the tied best actions $\mathcal{B}=\{i:y_i=\max_j y_j\}$, and the pairs with distinct labels $\mathcal{P}=\{(i,j):i<j,\ y_i\ne y_j\}$. The best-action objective assigns equal target probability to tied best actions, while the pairwise objective weights comparisons by their label differences:
\begin{equation}
\label{eq:app_ranking_objectives}
\begin{aligned}
\mathcal{L}_{\mathrm{best}}
&=-\frac{1}{|\mathcal{B}|}\sum_{i\in\mathcal{B}}\log p_i,\\
\mathcal{L}_{\mathrm{pair}}
&=\frac{1}{|\mathcal{P}|}\sum_{(i,j)\in\mathcal{P}}
w_{ij}\log\!\left(1+\exp\!\left[-\operatorname{sign}(y_i-y_j)(s_i-s_j)\right]\right),\\
w_{ij}&=\operatorname{clip}\!\left(\frac{|y_i-y_j|}{\sigma_y},\,0.25,\,4\right).
\end{aligned}
\end{equation}
Here, $\sigma_y$ is the standard deviation of training action labels, so larger label differences receive greater weight within fixed bounds. The two losses are averaged over eligible state groups and added with equal weights, $\mathcal{L}_{\mathrm{score}}=\mathcal{L}_{\mathrm{best}}+\mathcal{L}_{\mathrm{pair}}$. Groups with fewer than two actions or no distinguishable label differences are omitted. Small tolerances handle ties and near-zero label differences.

\textbf{Optimization and checkpoint selection.} The base training configuration uses AdamW with learning rate $3\times10^{-4}$, weight decay $0.04$, and $(\beta_1,\beta_2)=(0.9,0.95)$. Training uses 24 epochs, 32 state groups per batch, a 5\% warmup followed by cosine decay, and gradient clipping at norm 1.

Checkpoint selection uses validation normalized selection regret. For state $t$, let $i_t^\star=\arg\max_i s_i$ be the selected action. Its normalized regret is
\begin{equation}
\operatorname{Regret}_t=\frac{\max_i y_i-y_{i_t^\star}}{\max_i y_i-\min_i y_i}.
\label{eq:app_selection_regret}
\end{equation}
States with no label variation are excluded. We select the checkpoint with the lowest mean validation regret to prioritize useful action selection over property-regression accuracy.

\subsection{Generator Training Details}
\label{app:generator_details}

The context-conditioned generator is trained to complete local molecule-transfer
patterns. Its goal is to generate a target molecule conditioned on a small
context of related molecules, thereby learning chemically valid molecular edits and local transformation patterns within analogue series.

We train the generator in two stages. In the first stage, we initialize the
sequence-to-sequence backbone from a molecular Transformer pretrained on
pairwise molecular transformation data. This pretraining stage exposes the model
to a broad range of molecular structures and local edits, and provides a general
molecular generation prior. In the second stage, we insert a graph neural
network fusion module and fine-tune the resulting hybrid model on graph-based
subgraph completion tasks constructed from a ChEMBL-derived molecule-transfer
graph. Each training instance consists of a local molecular context as input and
one held-out related molecule as the target. This training scheme encourages the
generator to combine broad molecular-edit knowledge from pairwise pretraining
with local analogue-series pattern modeling from graph-based completion.

To construct the molecule-transfer graph, we first identify
matched molecular pairs from ChEMBL. We then connect molecules appearing in
these pairs into an undirected molecule-transfer graph. From this graph, we sample connected local
subgraphs as training examples. Each sampled subgraph contains six molecules:
five molecules are used as the input context, and the remaining molecule is held
out as the target. We construct 300k such subgraph-completion examples and train
the generator with supervised teacher forcing by maximizing the conditional
likelihood of the target molecule given the context.

Architecturally, we use T5Chem as the molecular sequence-to-sequence backbone.
A two-layer GNN fusion module is inserted after the encoder to aggregate
information across molecules in the input context before decoding the target
molecule. Unless otherwise stated, the context length is five molecules, the GNN
dropout rate is 0.1, and the GNN hidden dimension is set to match the hidden
dimension of the Transformer backbone. The model is optimized with AdamW using a
learning rate of \(10^{-4}\), weight decay of \(10^{-5}\), batch size 16, and
maximum input and target lengths of 96 and 64 tokens, respectively. We train for three epochs, selecting the checkpoint by lowest validation loss.

We consider three fine-tuning regimes to assess the contribution of different
model components: updating only the GNN module, updating the GNN and decoder,
and updating the full model. In all cases, optimization is performed only over
trainable parameters. Validation losses are averaged across workers, with a separate best checkpoint for each fine-tuning regime.

\subsection{Molecule Graph Expansion}
\label{app:graph_update}

This subsection details the acceptance and edge-construction rules for the graph
update in Eq.~\ref{eq:world_transition_main}. Generated candidates are canonicalized,
checked for validity, and deduplicated against the current graph and the generated
batch. Let $\mathcal{X}_t^{\mathrm{valid}}$ denote the resulting valid, unique
candidates not already present in $\mathcal{V}_t$. These candidates are evaluated
by the target property oracle regardless of MMP acceptance. The current graph $G_t$
contains the initial molecules and molecules accepted in earlier batches. We
accept a candidate if RDKit-based MMP matching identifies an edge to any molecule
in $\mathcal{V}_t$, then insert the accepted molecules and their verified MMP
edges using Eq.~\ref{eq:world_transition_main}. The accepted set and added edges are
\begin{equation}
\label{eq:mmp_accepted_candidates}
\begin{aligned}
\widetilde{\mathcal{X}}_t
&=\left\{
 x\in\mathcal{X}_t^{\mathrm{valid}}
 \;\middle|\;
 \exists u\in\mathcal{V}_t,\;x\sim u
\right\},\\
\mathcal{E}_t^{+}
&=\left\{
\{x,u\}
\;\middle|\;
x\in\widetilde{\mathcal{X}}_t,\;
u\in\mathcal{V}_t\cup\widetilde{\mathcal{X}}_t,\;
x\ne u,\;
x\sim u
\right\}.
\end{aligned}
\end{equation}
The acceptance decisions are completed against $G_t$ before the graph is updated.
Here, $\{x,u\}$ denotes an
undirected edge between molecules $x$ and $u$, with $x\sim u$ indicating an MMP
relation. In this set, $u$ may be either an existing molecule or another accepted
new molecule. Each accepted molecule therefore extends an existing MMP path from
the initial set. The expanded graph and new property observations guide subsequent anchor
selection, closing the loop between world model-guided generation and graph
expansion.

\subsection{Optimization Protocol}

At each iteration, \textsc{MolWorld} uses the molecule world model and property scorer to select anchor contexts from the current molecule-transfer graph, generates candidate molecules from each selected context, evaluates valid, deduplicated candidates with the target oracle, and expands the graph using RDKit-based MMP validation.
Repeated candidates do not consume additional oracle budget. Evaluations of new candidates count toward the budget regardless of whether they satisfy the MMP reachability criterion.
The updated molecule-transfer graph is then used for the next iteration.
For property optimization, the oracle budget is 10000 for ZINC250 and 1000 for PMV21. 
For docking-based optimization, the oracle budget is 1000. 
All experiments are repeated with 3 random seeds and conducted on a server equipped with an AMD EPYC 7413 CPU, 125 GiB system memory, and an NVIDIA RTX A6000 GPU with 48 GB of memory.

\section{Additional Experimental Details}
\subsection{Evaluation Scope}
\label{app:evaluation_scope}

\textbf{Oracle budgets.} The reported budgets count new oracle evaluations during optimization. For property optimization, the scorer is pretrained using target-property labels on ChEMBL-derived expansion outcomes and reused across searches for the same property. We report oracle query efficiency during optimization separately from this reusable training investment. The docking scorer uses only docking evaluations collected within the reported 1,000-call budget.

\textbf{Actionability.} Actionability here refers to structural-series actionability: verified MMP paths connect generated candidates to the initial series and provide structural references for subsequent optimization.

\subsection{Baseline Details}
\label{app:baseline_details}

We provide brief descriptions of the baseline methods used in our experiments. 
All baselines use the same oracle budgets during optimization and task settings as \textsc{MolWorld}.

We compare methods under fixed budgets of new oracle evaluations during optimization, with each method using its pretrained models and previously prepared resources. Several baselines incorporate prior information through pretrained molecular generators, learned latent representations, or property-informed fragment vocabularies. In particular, GenMol uses target-property scores computed before optimization to construct its initial fragment vocabulary. The reported budgets measure additional oracle evaluations during optimization; the cost of pretraining and offline resource construction is considered separately.

\paragraph{GraphGA.}
GraphGA~\citep{jensen2019graph} is a graph-based genetic algorithm for molecular optimization. 
It maintains a population of molecules and iteratively generates new candidates through molecular mutation and crossover operations. 
The population is updated according to oracle scores, allowing the search process to favor molecules with improved properties over successive generations.

\paragraph{MOLLEO.}
MOLLEO~\citep{wang2025efficient} is a recent molecule optimization method that combines evolutionary search with large language model-based molecular editing. 
We use its BioT5-based variant, which incorporates a model pretrained on molecular, biological, and textual data and subsequently fine-tuned for text-guided molecule generation. It uses LLM-guided mutation and crossover operations to propose chemically meaningful candidates, and then selects molecules based on oracle feedback.
This supports search under limited oracle budgets.

\paragraph{REINVENT.}
REINVENT~\citep{olivecrona2017molecular} is a reinforcement-learning-based molecular generation method built on a SMILES generative model. 
Starting from a pretrained prior model, it updates the generator using oracle-based rewards so that generation favors desired properties. The prior learns molecular sequence patterns before the target-specific optimization begins.

\paragraph{Augmented Memory.}
Augmented Memory~\citep{guo2023augmented} is a reinforcement-learning-based molecular optimization method that improves sample efficiency by reusing high-scoring molecules observed during optimization. 
It starts from a pretrained molecular prior and reuses stored oracle scores across augmented SMILES representations to improve learning without additional evaluations of the same molecules. The memory mechanism helps preserve promising molecular patterns and stabilizes oracle-guided generator updates.

\paragraph{GP-BO.}
GP-BO~\citep{tripp2021fresh} applies Bayesian optimization to molecular optimization. 
It trains a Gaussian process surrogate model to approximate the oracle function over molecular representations and uses the surrogate to guide the selection of promising candidate molecules.

\paragraph{NFBO.}
NFBO~\citep{lee2025latent} performs Bayesian optimization in the latent space of SeqFlow, an autoregressive normalizing flow for molecular sequences. The released SeqFlow checkpoint is pretrained on 1.27 million molecules from GuacaMol. Its encoding and decoding mappings are designed to eliminate the reconstruction gap between a molecule and its latent representation. Candidate sampling adjusts exploration at each token according to its importance, while oracle observations guide the latent-space search.

\paragraph{GenMol.}
GenMol~\citep{lee2025genmol} is a discrete diffusion model operating on Sequential Attachment-based Fragment Embedding (SAFE) sequences. Its generator is pretrained on the SAFE dataset containing molecules from ZINC and UniChem. It generates molecules through parallel denoising and supports molecular optimization through fragment remasking: selected fragments are regenerated while the remaining molecular context is retained. This allows local structural modifications to be explored using oracle feedback. Before optimization, its initial fragment vocabulary is ranked using the average target-property scores of molecules containing each fragment. These scores are computed offline and are separate from the oracle calls counted during optimization.

\paragraph{RetMol.}
RetMol~\citep{wang2022retrieval} guides a pretrained molecular generator with retrieved exemplar molecules that partially or fully satisfy the desired property criteria. Its retrieval fusion module is trained using a self-supervised molecular similarity objective, while the pretrained encoder and decoder remain frozen. It fuses representations of the input molecule and retrieved exemplars to condition generation. Iterative refinement updates molecules and the retrieval database, using new discoveries to guide proposals without task-specific fine-tuning.

\paragraph{InVirtuoGen.}
InVirtuoGen~\citep{kaech2026refine} is a discrete flow model for fragmented SMILES that refines sequences starting from a uniform token distribution. Optimization starts from a pretrained generative checkpoint. For property optimization, it combines genetic search with Proximal Property Optimization, a fine-tuning procedure adapted to discrete flows. Oracle feedback guides both candidate selection and updates to the generative model.

\subsection{Docking Protocol}
\label{app:docking_protocol}

Task 2 evaluates generated molecules on DRD3, EGFR, and A2AR using TDC docking oracles through the PyScreener AutoDock Vina backend. The corresponding receptor structures are 3PBL, 2RGP, and 3EML. For each molecule, the oracle returns the best Vina binding affinity among the generated poses, with lower scores indicating stronger predicted binding.

PyScreener converts the prepared receptor PDB files to PDBQT format using the \texttt{prepare\_receptor} utility. For ligand preparation, generated SMILES strings are parsed with RDKit, hydrogens are added, and 3D conformers are generated using distance-geometry embedding and minimized with the MMFF force field. The ligands are then processed with OpenBabel/Pybel, assigned Gasteiger charges, and written in PDBQT format.

The docking box centers and sizes follow the TDC metadata (Table~\ref{tab:docking_protocol_settings}). Vina uses an exhaustiveness of 8, up to 9 output poses, an energy range of 3.0 kcal/mol, and 4 CPU threads per docking call. The Vina random seed is not fixed.

Docking AUC is computed over the first 1,000 oracle calls using $\bar{s}_{\mathrm{dock}}=\min(0,s_{\mathrm{dock}})$. For all compared methods, molecule eligibility follows the batch-wise reachability criterion defined in the experimental setup. Evaluations remain on the oracle-call axis regardless of molecule eligibility. When batch records are unavailable, molecules are processed individually in the saved oracle-call order. The running top-$k$ score sums the $k$ lowest eligible scores and divides by $\min(k,t)$, with unfilled positions assigned zero. AUC@10 and AUC@100 average these running scores across calls, so lower values are better. Within the running top-$k$ set, duplicate molecules occupy one position, using their best eligible observed score; failed calls retain their budget positions. Only scores available during the original run are included. Iso is the fraction of valid, scored unique generated molecules that are not accepted under the same reachability criterion.

\begin{table}[t]
\centering
\caption{Task 2 docking settings using TDC boxes (coordinates and sizes in \AA).}
\label{tab:docking_protocol_settings}
\small
\begin{tabular}{llllll}
\toprule
Target & PDB ID & Box center $(x,y,z)$ & Box size $(x,y,z)$ & Exhaustiveness & Poses \\
\midrule
DRD3 & 3PBL & $(9,\ 22.5,\ 26)$ & $(15,\ 15,\ 15)$ & 8 & 9 \\
EGFR & 2RGP & $(16.2921,\ 34.8708,\ 92.0353)$ & $(15,\ 15,\ 15)$ & 8 & 9 \\
A2AR & 3EML & $(-9.0636,\ -7.1446,\ 55.8626)$ & $(15,\ 15,\ 15)$ & 8 & 9 \\
\bottomrule
\end{tabular}
\end{table}

\subsection{Additional Property Optimization Results}
\label{app:task1_std}

This section provides detailed property optimization results with standard deviations.
Tables~\ref{tab:zinc_results_reachable_std} and~\ref{tab:pmv21_results_reachable_std} report the MMP-constrained results on ZINC250 and PMV21, respectively, using the batch-wise reachability criterion defined in the experimental setup.
ZINC250 results are evaluated with 10,000 oracle calls, while PMV21 results are evaluated with 1,000 oracle calls.
Results are reported as mean $\pm$ sample standard deviation across seeds.
None denotes the absence of generated molecules satisfying the MMP connectivity constraint.

\begin{table*}[t]
\centering
\scriptsize
\setlength{\tabcolsep}{3pt}
\renewcommand{\arraystretch}{1.05}
\caption{MMP-constrained property optimization on ZINC250. Results are reported as mean $\pm$ sample standard deviation. Iso denotes the fraction of generated molecules that do not connect to the initial graph under the reachability criterion.}
\label{tab:zinc_results_reachable_std}
\resizebox{\textwidth}{!}{%
\begin{tabular}{l*{6}{c}}
\toprule
Method & \multicolumn{3}{c}{GSK3$\beta$} & \multicolumn{3}{c}{QED} \\
\cmidrule(lr){2-4}\cmidrule(lr){5-7}
 & AUC@10$\uparrow$ & AUC@100$\uparrow$ & Iso$\downarrow$ & AUC@10$\uparrow$ & AUC@100$\uparrow$ & Iso$\downarrow$ \\
\midrule
GraphGA & 0.3963 $\pm$ 0.3076 & 0.2589 $\pm$ 0.3407 & 0.8792 $\pm$ 0.1763 & 0.9304 $\pm$ 0.0046 & 0.8844 $\pm$ 0.0095 & 0.8092 $\pm$ 0.0208 \\
REINVENT & 0.6564 $\pm$ 0.2467 & 0.5338 $\pm$ 0.2734 & 0.6689 $\pm$ 0.1972 & 0.9392 $\pm$ 0.0007 & 0.9062 $\pm$ 0.0025 & 0.6274 $\pm$ 0.0483 \\
Aug-Mem & 0.6175 $\pm$ 0.2580 & 0.4400 $\pm$ 0.3527 & 0.9091 $\pm$ 0.0676 & 0.9366 $\pm$ 0.0011 & 0.8931 $\pm$ 0.0047 & 0.7300 $\pm$ 0.0421 \\
GP-BO & 0.8195 $\pm$ 0.1652 & 0.7198 $\pm$ 0.2186 & 0.6106 $\pm$ 0.3663 & 0.9231 $\pm$ 0.0056 & 0.8745 $\pm$ 0.0060 & 0.9330 $\pm$ 0.0096 \\
MOLLEO & 0.8904 $\pm$ 0.0425 & \underline{0.8090 $\pm$ 0.0503} & 0.4642 $\pm$ 0.1389 & 0.9299 $\pm$ 0.0087 & 0.8751 $\pm$ 0.0335 & 0.7956 $\pm$ 0.0287 \\
NFBO & 0.8316 $\pm$ 0.0235 & 0.6026 $\pm$ 0.0187 & 0.9623 $\pm$ 0.0082 & 0.9384 $\pm$ 0.0014 & 0.8856 $\pm$ 0.0063 & 0.7273 $\pm$ 0.0320 \\
GenMol & 0.3360 $\pm$ 0.2550 & 0.1083 $\pm$ 0.1172 & 0.9901 $\pm$ 0.0094 & 0.9103 $\pm$ 0.0093 & 0.8266 $\pm$ 0.0057 & 0.8720 $\pm$ 0.0089 \\
RetMol & 0.5112 $\pm$ 0.4455 & 0.4687 $\pm$ 0.4099 & \underline{0.4019 $\pm$ 0.1239} & \underline{0.9457 $\pm$ 0.0015} & \underline{0.9396 $\pm$ 0.0035} & \underline{0.3427 $\pm$ 0.2195} \\
InVirtuoGen & \underline{0.9012 $\pm$ 0.0310} & 0.7397 $\pm$ 0.0522 & 0.7102 $\pm$ 0.0737 & 0.9288 $\pm$ 0.0136 & 0.8646 $\pm$ 0.0452 & 0.8655 $\pm$ 0.0865 \\
MolWorld & \textbf{0.9576 $\pm$ 0.0216} & \textbf{0.9044 $\pm$0.0385} & \textbf{0.0460 $\pm$ 0.0043} & \textbf{0.9481 $\pm$ 0.0001} & \textbf{0.9470 $\pm$ 0.0005} & \textbf{0.0438 $\pm$ 0.0066} \\
\midrule
Method & \multicolumn{3}{c}{JNK3} & \multicolumn{3}{c}{DRD2} \\
\cmidrule(lr){2-4}\cmidrule(lr){5-7}
 & AUC@10$\uparrow$ & AUC@100$\uparrow$ & Iso$\downarrow$ & AUC@10$\uparrow$ & AUC@100$\uparrow$ & Iso$\downarrow$ \\
\midrule
GraphGA & 0.5056 $\pm$ 0.1334 & 0.4081 $\pm$ 0.1085 & 0.8233 $\pm$ 0.1054 & 0.6163 $\pm$ 0.3437 & 0.3150 $\pm$ 0.3110 & 0.8772 $\pm$ 0.0408 \\
REINVENT & 0.7574 $\pm$ 0.0303 & 0.7097 $\pm$ 0.0320 & 0.3953 $\pm$ 0.0572 & 0.8903 $\pm$ 0.0134 & 0.8381 $\pm$ 0.0113 & 0.5380 $\pm$ 0.1838 \\
Aug-Mem & 0.7201 $\pm$ 0.0944 & 0.6445 $\pm$ 0.0961 & 0.7495 $\pm$ 0.1489 & 0.8552 $\pm$ 0.0375 & 0.7426 $\pm$ 0.0441 & 0.7895 $\pm$ 0.0800 \\
GP-BO & 0.3509 $\pm$ 0.1783 & 0.3050 $\pm$ 0.1919 & 0.3230 $\pm$ 0.2317 & 0.9083 $\pm$ 0.0621 & 0.8336 $\pm$ 0.0679 & \underline{0.3163 $\pm$ 0.2214} \\
MOLLEO & \underline{0.7759 $\pm$ 0.0471} & \underline{0.7106 $\pm$ 0.0519} & 0.6214 $\pm$ 0.0395 & 0.7474 $\pm$ 0.3493 & 0.3679 $\pm$ 0.4014 & 0.8543 $\pm$ 0.0849 \\
NFBO & 0.6459 $\pm$ 0.0257 & 0.5559 $\pm$ 0.0379 & 0.6462 $\pm$ 0.1412 & 0.9179 $\pm$ 0.0540 & 0.7823 $\pm$ 0.0266 & 0.7489 $\pm$ 0.0770 \\
GenMol & 0.0003 $\pm$ 0.0006 & 0.0000 $\pm$ 0.0001 & 1.0000 $\pm$ 0.0001 & \underline{0.9860 $\pm$ 0.0027} & \underline{0.9522 $\pm$ 0.0122} & 0.4267 $\pm$ 0.0266 \\
RetMol & 0.5422 $\pm$ 0.1701 & 0.4656 $\pm$ 0.1678 & \underline{0.1996 $\pm$ 0.0321} & 0.9484 $\pm$ 0.0359 & 0.8536 $\pm$ 0.1695 & 0.4884 $\pm$ 0.2729 \\
InVirtuoGen & 0.6833 $\pm$ 0.2557 & 0.5412 $\pm$ 0.3364 & 0.7918 $\pm$ 0.1663 & 0.6805 $\pm$ 0.2432 & 0.1792 $\pm$ 0.1431 & 0.7820 $\pm$ 0.0600 \\
MolWorld & \textbf{0.8225 $\pm$ 0.1293} & \textbf{0.7770 $\pm$ 0.1229} & \textbf{0.1105 $\pm$ 0.0951} & \textbf{0.9943 $\pm$ 0.0010} & \textbf{0.9819 $\pm$ 0.0014} & \textbf{0.0202 $\pm$ 0.0020} \\
\bottomrule
\end{tabular}%
}
\end{table*}

\begin{table*}[t]
\centering
\scriptsize
\setlength{\tabcolsep}{3pt}
\renewcommand{\arraystretch}{1.05}
\caption{MMP-constrained property optimization on PMV21. Results are reported as mean $\pm$ sample standard deviation. Iso denotes the fraction of generated molecules that do not connect to the initial graph under the reachability criterion.}
\label{tab:pmv21_results_reachable_std}
\resizebox{\textwidth}{!}{%
\begin{tabular}{l*{6}{c}}
\toprule
Method & \multicolumn{3}{c}{GSK3$\beta$} & \multicolumn{3}{c}{QED} \\
\cmidrule(lr){2-4}\cmidrule(lr){5-7}
 & AUC@10$\uparrow$ & AUC@100$\uparrow$ & Iso$\downarrow$ & AUC@10$\uparrow$ & AUC@100$\uparrow$ & Iso$\downarrow$ \\
\midrule
GraphGA & 0.2751 $\pm$ 0.0217 & 0.1856 $\pm$ 0.0134 & \underline{0.6057 $\pm$ 0.0558} & 0.7286 $\pm$ 0.0362 & 0.5408 $\pm$ 0.0238 & 0.8007 $\pm$ 0.0205 \\
REINVENT & None & None & 1.0000 $\pm$ 0.0000 & None & None & 1.0000 $\pm$ 0.0000 \\
Aug-Mem & None & None & 1.0000 $\pm$ 0.0000 & None & None & 1.0000 $\pm$ 0.0000 \\
GP-BO & 0.2426 $\pm$ 0.0046 & 0.1512 $\pm$ 0.0015 & 0.6597 $\pm$ 0.0040 & 0.6908 $\pm$ 0.0516 & 0.5429 $\pm$ 0.0230 & \underline{0.6237 $\pm$ 0.0393} \\
MOLLEO & 0.3008 $\pm$ 0.0162 & \underline{0.1936 $\pm$ 0.0134} & 0.6603 $\pm$ 0.0902 & 0.7880 $\pm$ 0.0194 & \underline{0.5649 $\pm$ 0.0233} & 0.7903 $\pm$ 0.0202 \\
NFBO & 0.2733 $\pm$ 0.0484 & 0.1082 $\pm$ 0.0433 & 0.8463 $\pm$ 0.0764 & 0.6115 $\pm$ 0.1316 & 0.3218 $\pm$ 0.1050 & 0.8445 $\pm$ 0.0683 \\
GenMol & \underline{0.3138 $\pm$ 0.0394} & 0.0538 $\pm$ 0.0095 & 0.9810 $\pm$ 0.0017 & \underline{0.8010 $\pm$ 0.0482} & 0.3158 $\pm$ 0.1516 & 0.9477 $\pm$ 0.0294 \\
RetMol & 0.2037 $\pm$ 0.0291 & 0.0866 $\pm$ 0.0168 & 0.8917 $\pm$ 0.0080 & 0.6290 $\pm$ 0.0336 & 0.4175 $\pm$ 0.0152 & 0.9000 $\pm$ 0.0020 \\
InVirtuoGen & 0.2287 $\pm$ 0.0069 & 0.1284 $\pm$ 0.0031 & 0.9000 $\pm$ 0.0020 & 0.6113 $\pm$ 0.0041 & 0.4287 $\pm$ 0.0038 & 0.8973 $\pm$ 0.0012 \\
MolWorld & \textbf{0.4225 $\pm$ 0.0159} & \textbf{0.3490 $\pm$ 0.0220} & \textbf{0.0493 $\pm$ 0.0060} & \textbf{0.8892 $\pm$ 0.0119} & \textbf{0.8194 $\pm$ 0.0098} & \textbf{0.0744 $\pm$ 0.0359} \\
\midrule
Method & \multicolumn{3}{c}{JNK3} & \multicolumn{3}{c}{DRD2} \\
\cmidrule(lr){2-4}\cmidrule(lr){5-7}
 & AUC@10$\uparrow$ & AUC@100$\uparrow$ & Iso$\downarrow$ & AUC@10$\uparrow$ & AUC@100$\uparrow$ & Iso$\downarrow$ \\
\midrule
GraphGA & 0.1490 $\pm$ 0.0148 & 0.0982 $\pm$ 0.0074 & 0.5527 $\pm$ 0.0389 & 0.7722 $\pm$ 0.0023 & 0.4403 $\pm$ 0.0225 & 0.6247 $\pm$ 0.0567 \\
REINVENT & None & None & 1.0000 $\pm$ 0.0000 & None & None & 1.0000 $\pm$ 0.0000 \\
Aug-Mem & None & None & 1.0000 $\pm$ 0.0000 & None & None & 1.0000 $\pm$ 0.0000 \\
GP-BO & 0.1364 $\pm$ 0.0159 & 0.0858 $\pm$ 0.0050 & 0.6420 $\pm$ 0.0272 & 0.7674 $\pm$ 0.0632 & 0.4145 $\pm$ 0.0642 & \underline{0.5930 $\pm$ 0.1127} \\
MOLLEO & 0.1744 $\pm$ 0.0102 & \underline{0.1114 $\pm$ 0.0048} & \underline{0.5347 $\pm$ 0.0331} & 0.8019 $\pm$ 0.0173 & 0.4521 $\pm$ 0.0369 & 0.7047 $\pm$ 0.0400 \\
NFBO & 0.1662 $\pm$ 0.0543 & 0.0951 $\pm$ 0.0639 & 0.6734 $\pm$ 0.2556 & 0.6783 $\pm$ 0.1008 & 0.3092 $\pm$ 0.2019 & 0.7897 $\pm$ 0.0709 \\
GenMol & \underline{0.2315 $\pm$ 0.0698} & 0.0427 $\pm$ 0.0150 & 0.9717 $\pm$ 0.0040 & \underline{0.8659 $\pm$ 0.0313} & 0.3765 $\pm$ 0.0735 & 0.9123 $\pm$ 0.0244 \\
RetMol & 0.1122 $\pm$ 0.0083 & 0.0529 $\pm$ 0.0047 & 0.8930 $\pm$ 0.0030 & 0.7774 $\pm$ 0.1405 & \underline{0.5135 $\pm$ 0.2820} & 0.6157 $\pm$ 0.2333 \\
InVirtuoGen & 0.1321 $\pm$ 0.0025 & 0.0741 $\pm$ 0.0014 & 0.9000 $\pm$ 0.0010 & 0.4878 $\pm$ 0.0693 & 0.2112 $\pm$ 0.0356 & 0.7478 $\pm$ 0.0072 \\
MolWorld & \textbf{0.2710 $\pm$ 0.0286} & \textbf{0.2175 $\pm$ 0.0224} & \textbf{0.0437 $\pm$ 0.0090} & \textbf{0.9018 $\pm$ 0.0063} & \textbf{0.8205 $\pm$ 0.0047} & \textbf{0.0990 $\pm$ 0.0249} \\
\bottomrule
\end{tabular}%
}
\end{table*}

InVirtuoGen exhibits substantial variation on ZINC--JNK3, with AUC@10 values of 0.8513, 0.3890, and 0.8096 across three seeds. The lowest-AUC run eventually achieves a final top-10 mean score of 0.917 among reachable molecules, but its first reachable molecule scoring at least 0.8 appears only after 6,637 oracle evaluations, compared with 1,185 and 1,302 evaluations in the other runs. This delayed discovery substantially reduces its AUC despite strong final scores. In another seed, InVirtuoGen achieves a maximum reachable score of 0.98, exceeding the maximum of 0.96 observed across the three \textsc{MolWorld} runs. These results illustrate the distinction between finding an exceptionally high-scoring molecule and consistently discovering high-scoring, reachable molecules early within a fixed oracle budget.

\textbf{Training-molecule overlap.} We examine exact molecular overlap between the molecules used to construct property-scorer training labels and the initial and generated molecules in \textsc{MolWorld}. Molecules are matched using canonical isomeric SMILES. The initial PMV21 graph contains no overlapping molecules among 722 molecules, while the ZINC250 graph contains 341 among 109,500 (0.3114\%). Generated-molecule overlap is computed over the union of unique molecules across three seeds. Final Top-10 and Top-100 overlap counts are summed across the three seeds, considering only generated molecules that satisfy the MMP reachability criterion.

To evaluate the contribution of overlapping molecules to optimization scores, we set their score contributions to zero and recompute AUC while preserving oracle-call positions, budgets, and reachability labels. We report the mean difference across seeds, defined as AUC after exclusion minus AUC before exclusion (Table~\ref{tab:property_training_overlap}). Final Top-10 and Top-100 overlap describe the molecules retained at the end of optimization, whereas AUC changes also reflect overlapping molecules that entered the top-ranked sets earlier in the search.

\begin{table}[htbp]
\centering
\small
\setlength{\tabcolsep}{3pt}
\renewcommand{\arraystretch}{1.08}
\caption{Training-molecule overlap and AUC sensitivity of \textsc{MolWorld}. Input and generated overlap are percentages. Final Top-10 and Top-100 overlap are molecule counts summed across three seeds. $\Delta$AUC reports the mean change after excluding overlapping score contributions.}
\label{tab:property_training_overlap}
\begin{tabular}{llrrrrrr}
\toprule
Dataset & Property & \shortstack{Input\\overlap (\%)} & \shortstack{Generated\\overlap (\%)} & \shortstack{Final Top-10\\overlap} & \shortstack{Final Top-100\\overlap} & $\Delta$AUC@10 & $\Delta$AUC@100 \\
\midrule
\multirow{4}{*}{ZINC250} & GSK3$\beta$ & \multirow{4}{*}{0.3114} & 0.0156 & 0 & 0 & $-0.000016$ & $-0.000075$ \\
 & QED & & 3.5317 & 0 & 1 & 0.000000 & $-0.000656$ \\
 & JNK3 & & 0.3385 & 0 & 0 & $-0.000469$ & $-0.000344$ \\
 & DRD2 & & 0.0489 & 0 & 0 & 0.000000 & $-0.000041$ \\
\midrule
\multirow{4}{*}{PMV21} & GSK3$\beta$ & \multirow{4}{*}{0.0000} & 0.0000 & 0 & 0 & 0.000000 & 0.000000 \\
 & QED & & 0.0000 & 0 & 0 & 0.000000 & 0.000000 \\
 & JNK3 & & 0.0000 & 0 & 0 & 0.000000 & 0.000000 \\
 & DRD2 & & 0.0000 & 0 & 0 & 0.000000 & 0.000000 \\
\bottomrule
\end{tabular}
\end{table}

Across the analyzed trajectories, none of the final Top-10 molecules overlaps with the property-scorer training molecules. The final Top-100 sets contain only one overlapping molecule, found in the ZINC250 QED task. Excluding overlapping score contributions reduces mean AUC@10 by at most 0.000469 and mean AUC@100 by at most 0.000656, indicating that these molecules contribute little directly to the AUC scores in this analysis.

\textbf{Training-scaffold overlap.} We further compare generated molecules with the property-scorer label training set using Bemis--Murcko scaffolds computed with RDKit, ignoring stereochemistry. A generated molecule counts as overlapping only if its scaffold occurs in the training set but is absent from the initial input graph. Scaffolds already present in the input graph are excluded from the overlap counts, while the denominators remain unchanged. We consider only MMP-reachable generated molecules. The output percentage is computed over the union of unique molecules across three seeds; final Top-10 and Top-100 counts sum the ranked entries across seeds.

\begin{table}[htbp]
\centering
\small
\setlength{\tabcolsep}{7pt}
\renewcommand{\arraystretch}{1.08}
\caption{Scaffold overlap with the property-scorer label training set after excluding scaffolds present in the initial graph. Generated overlap is the percentage of output molecules satisfying this condition. Top-10 and Top-100 entries report overlapping entries out of the total across three seeds.}
\label{tab:property_training_scaffold_overlap}
\begin{tabular}{llrrr}
\toprule
Dataset & Property & Generated overlap (\%) & Top-10 overlap & Top-100 overlap \\
\midrule
\multirow{4}{*}{ZINC250} & GSK3$\beta$ & 1.60 & 0/30 & 0/300 \\
 & QED & 9.66 & 0/30 & 15/300 \\
 & JNK3 & 49.96 & 20/30 & 198/300 \\
 & DRD2 & 39.23 & 18/30 & 158/300 \\
\midrule
\multirow{4}{*}{PMV21} & GSK3$\beta$ & 0.00 & 0/30 & 0/300 \\
 & QED & 0.05 & 0/30 & 0/300 \\
 & JNK3 & 0.00 & 0/30 & 0/300 \\
 & DRD2 & 0.00 & 0/30 & 0/300 \\
\bottomrule
\end{tabular}
\end{table}

Table~\ref{tab:property_training_scaffold_overlap} shows that overlap varies across tasks, with higher rates on ZINC250 JNK3 and DRD2. Our objective is to improve molecular properties while preserving MMP reachability from the initial set; scaffold novelty is not an optimization requirement. Sharing a scaffold with a training molecule does not imply reproducing that molecule, so scaffold overlap complements the exact-molecule analysis above. On PMV21, the initial graph shares no scaffolds with the property-scorer label training set, and generated-molecule overlap is zero for GSK3$\beta$, JNK3, and DRD2, including their final Top-10 and Top-100 sets. Nevertheless, \textsc{MolWorld} achieves the highest AUC@10 and AUC@100 among the compared methods on each of these three tasks (Table~\ref{tab:property_optimization_main}). Strong optimization performance is therefore also observed on tasks with no measured scaffold overlap with this training set.

\subsection{Optimization Dynamics}
\label{sec:convergence}

\begin{figure}[h]             
  \centering                  
  \includegraphics[width=0.9\textwidth]{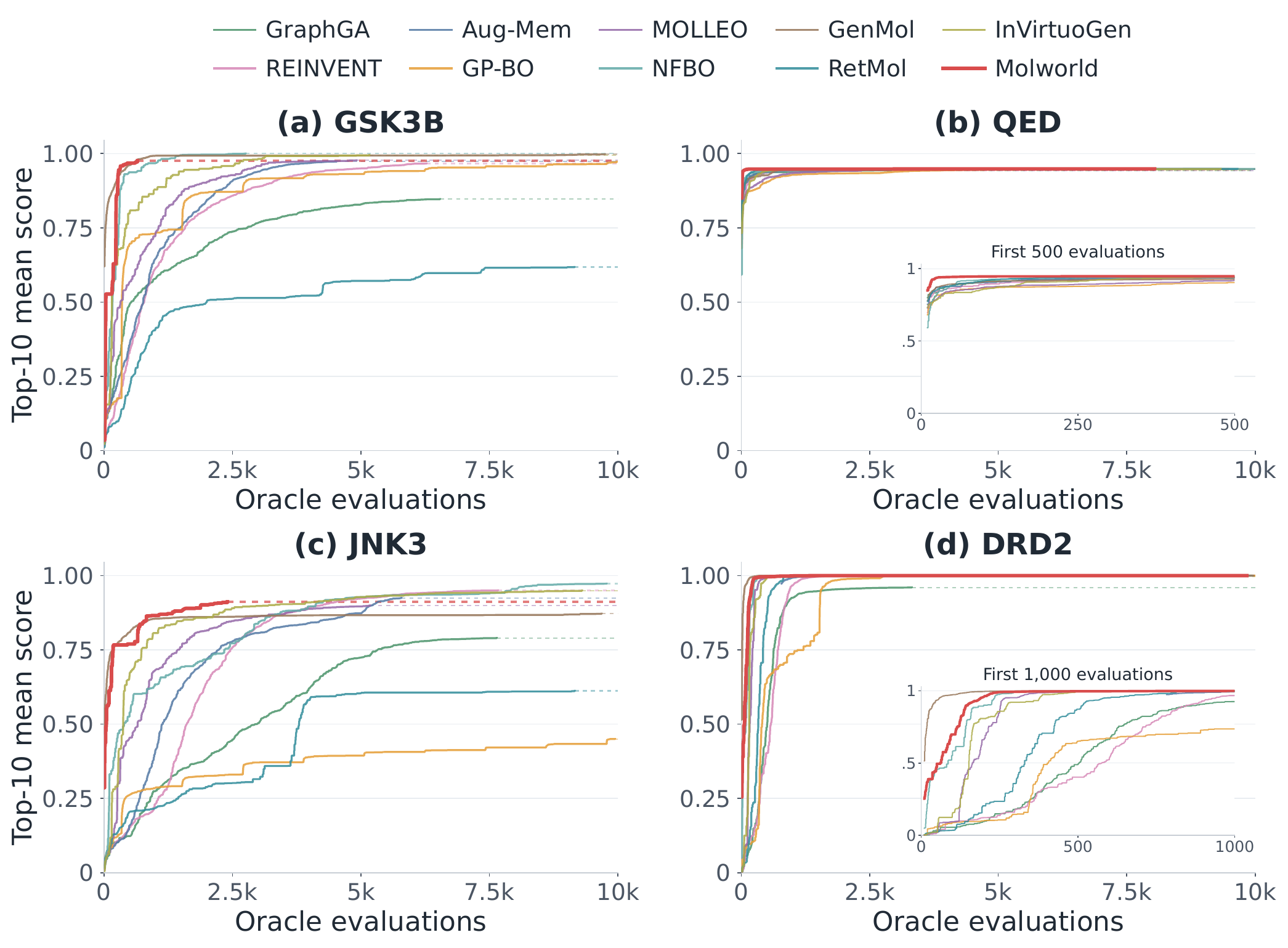} 
  \caption{Optimization dynamics on ZINC250. Top-10 running average oracle score over all valid scored molecules as a function of oracle calls across four tasks.}
  \label{fig:conv_zinc}         
\end{figure}

We analyze the optimization dynamics of different methods by plotting the top-10 running average oracle score as a function of oracle calls. 
Figure~\ref{fig:conv_zinc} shows how efficiently each method discovers high-quality molecules under the same oracle budget.

\textsc{MolWorld} shows faster improvement on more challenging tasks such as GSK3$\beta$ and JNK3, reaching strong top-10 performance earlier than most baselines.
On relatively easier tasks such as QED and DRD2, most methods improve quickly, while \textsc{MolWorld} remains competitive throughout the search process.
These results suggest that \textsc{MolWorld} efficiently guides search toward high-quality regions under fixed oracle budgets.


\subsection{Additional Generated Molecule Distribution Results}
\label{app:graph_extension}

\begin{figure}[h]
  \centering
  \includegraphics[width=0.85\textwidth]{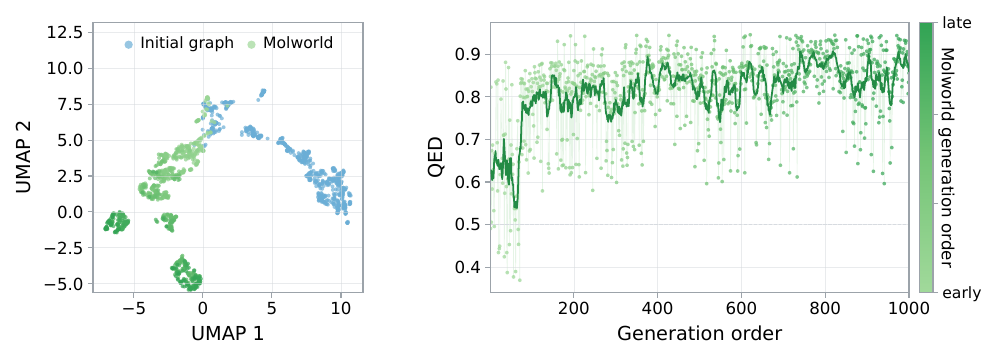}
  \caption{Generated molecule distribution and oracle-score progression of \textsc{MolWorld} on QED.}
  \label{fig:qed_extension}
\end{figure}

\begin{figure}[h]
  \centering
  \includegraphics[width=0.85\textwidth]{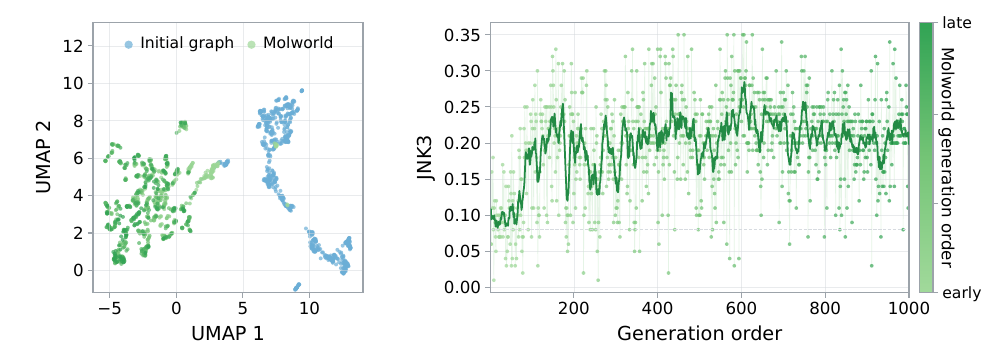}
  \caption{Generated molecule distribution and oracle-score progression of \textsc{MolWorld} on JNK3.}
  \label{fig:jnk3_extension}
\end{figure}

\begin{figure}[h]
  \centering
  \includegraphics[width=0.85\textwidth]{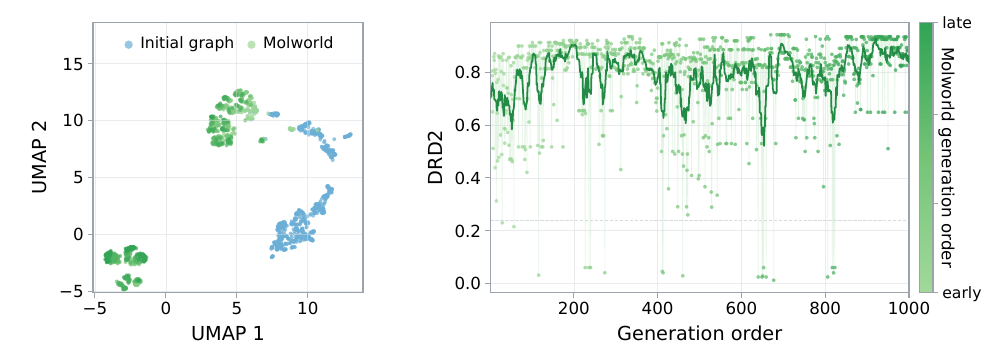}
  \caption{Generated molecule distribution and oracle-score progression of \textsc{MolWorld} on DRD2.}
  \label{fig:drd2_extension}
\end{figure}

We provide additional generated molecule distribution visualizations on QED, JNK3, and DRD2 in Figures~\ref{fig:qed_extension}, \ref{fig:jnk3_extension}, and~\ref{fig:drd2_extension}. 
For each task, the left panel shows the UMAP projection of initial and generated molecules, and the right panel shows oracle-score progression over generation order with a 10-step moving-average curve. 
Consistent with the GSK3$\beta$ result in the main text, \textsc{MolWorld} expands from the initial molecule-transfer graph toward higher-scoring regions.

\FloatBarrier
\subsection{Anchor Context Evolution}
\label{sec:anchor_evolution}

\begin{figure}[!htbp]
  \centering                  
  \includegraphics[width=0.9\textwidth]{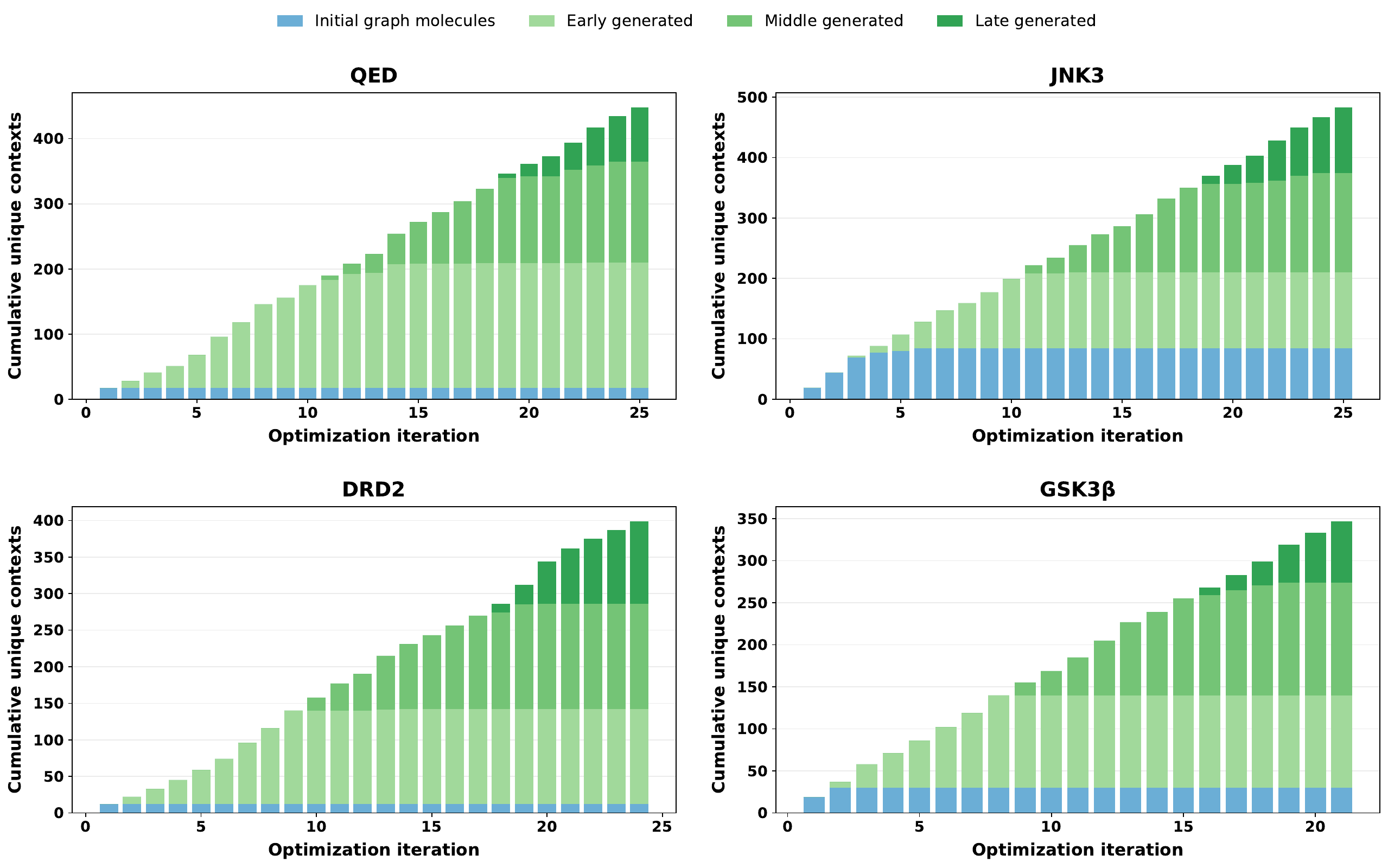} 
  \caption{Evolution of anchor context sources across iterations, shown as cumulative counts. Blue denotes initial-graph molecules; light, medium, and dark green denote molecules from early, middle, and late stages.}
  \label{fig:ap2}          
\end{figure}

We further analyze how the selected anchor contexts evolve during optimization, as shown in Figure~\ref{fig:ap2}. 
Each anchor context contains five molecules that define the local generation region. 
Blue bars denote initial-graph molecules; light, medium, and dark green bars denote molecules generated in early, middle, and late stages, respectively.

Across all tasks, anchor contexts quickly shift from initial molecules to newly generated molecules after the early iterations. 
The number of initial molecules used in anchor contexts increases only at the beginning and then stabilizes, while generated molecules become the dominant source of later anchor contexts. 
This confirms the self-expanding behavior of \textsc{MolWorld}: newly generated molecules are incorporated into the evolving molecule-transfer graph and reused to guide subsequent generation. 
As a result, \textsc{MolWorld} can progressively build upon its own generated molecules rather than repeatedly sampling from the initial graph.

\end{document}